\documentclass[letterpaper]{article} 
\usepackage{aaai2026}  
\usepackage{times}  
\usepackage{helvet}  
\usepackage{courier}  
\usepackage[hyphens]{url}  
\usepackage{graphicx} 
\usepackage{natbib}  
\usepackage{caption} 
\usepackage{url}            
\usepackage{booktabs}       
\usepackage{amsfonts}       
\usepackage{nicefrac}       
\usepackage{microtype}      
\usepackage{multirow}
\usepackage[dvipsnames]{xcolor}
\usepackage{amssymb,amsmath,amsthm,mathtools,tikz,pgfplots,sansmath}
\usetikzlibrary{automata,calc,shapes,shapes.arrows,shapes.multipart,chains,arrows,%
  positioning,intersections,matrix,fit,backgrounds,arrows.meta,%
  circuits.logic.US,shapes.gates.logic.US}
\pgfplotsset{compat=newest}
\usepackage[group-separator={,}]{siunitx}
\usepackage{fontawesome5}
\usepackage{enumitem}
\usepackage[most]{tcolorbox}
\usepackage{listings}

\usepackage{caption}
\usepackage{subcaption}
\usepackage{adjustbox}

\usepackage[linesnumbered, ruled, vlined]{algorithm2e}
\SetEndCharOfAlgoLine{} 

\theoremstyle{plain}

\theoremstyle{definition}

\definecolor{lbrown}{HTML}{808080} 

\newtcolorbox{mybox}[2][]{%
  enhanced,
  boxsep=3pt,
  arc=1.25ex,
  colback=white,
  colframe=lbrown,
  boxrule=3pt,
  leftrule=18pt,
  overlay unbroken and first ={%
    \node[rotate=90,
          minimum width=1cm,
          anchor=south,
          font=\Large\sffamily\bfseries,
          yshift=-18pt,
          white]
    at (frame.west) {#2};
  }
}

\usepackage{listings}

 \definecolor{dkgreen}{rgb}{0,0.6,0}
 \definecolor{gray}{rgb}{0.5,0.5,0.5}
 \definecolor{mauve}{rgb}{0.58,0,0.82}

\usepackage{newfloat}
\usepackage{listings}
\DeclareCaptionStyle{ruled}{labelfont=normalfont,labelsep=colon,strut=off} 
\nocopyright

\title{Explaining Hitori Puzzles: Neurosymbolic Proof
Staging for Sequential Decisions}
\author{
    Maria Leonor Pacheco\equalcontrib, Fabio Somenzi\equalcontrib, Dananjay Srinivas\equalcontrib, Ashutosh Trivedi\equalcontrib\\
}
\affiliations{
    University of Colorado Boulder\\

    \{maria.pacheco, fabio, dananjay.srinivas, ashutosh.trivedi\}@colorado.edu
}

\usepackage{bibentry}

\begin{document}

\maketitle

\begin{abstract}

  We propose a neurosymbolic approach to the explanation of complex sequences of decisions that combines the strengths of decision procedures and Large Language Models (LLMs).  We demonstrate this approach by producing explanations for the solutions of Hitori puzzles.  The rules of Hitori include local constraints that are effectively explained by short resolution proofs.  However, they also include a connectivity constraint that is more suitable for visual explanations.  Hence, Hitori provides an excellent testing ground for a flexible combination of SAT solvers and LLMs.  We have implemented a tool that assists humans in solving Hitori puzzles, and we present experimental evidence of its effectiveness.
\end{abstract}
\section{Introduction}
\label{sec:intro}

Large Language Models (LLMs) are increasingly being used to support individual and specialized decision making in a wide range of domains~\cite{Chkirbene2024LargeLM,Kammere087469,kim2025toolsunderstandingheavyusers}. This success can be partially attributed to their ability to adapt to new tasks without ad hoc training by conditioning on the local context provided during inference~\cite{NEURIPS2020_1457c0d6,NEURIPS2020_6b493230}. 
While quick decisions are often useful, explaining the underlying rationale behind these decisions is essential for successful human-AI collaboration. However, LLMs have been consistently shown to struggle to produce consistent and faithful reasoning steps for complex tasks~\cite{turpin2023language,chen2024two}. Multistep and complex decisions require logically consistent and structured reasoning of the kind provided by decision procedures such as SAT solvers. When the problem is properly posed, SAT solvers not only discover solutions but can also produce formal proofs to justify their reasoning. However, these proofs are often lengthy and cumbersome to parse, even for domain experts with advanced knowledge of problem encoding and theorem proving.
To address this challenge, we present a neuro-symbolic approach that leverages LLMs to explain the proofs produced by SAT solvers in a context-dependent and human-adaptable manner, facilitating trustworthy, clearer, and more accessible explanations.

A natural testing ground for this approach arises in the context of logic-based puzzles such as Hitori and Sudoku, which are enjoyed by millions of people worldwide. In designing such puzzles, the setter often envisions a logical path of discovery, and solving the puzzle involves reconstructing this path. The joy lies not in arriving at a solution, but in uncovering a clear, step-by-step explanation that leads to it.
SAT/SMT solvers, such as Z3, are frequently employed to solve these puzzles and can also generate formal proofs of correctness. However, due to the sequential nature of the reasoning required, structuring intermediate subgoals to minimize cognitive load becomes a key challenge. We refer to this challenge as \emph{proof staging}.
In this paper, we introduce a framework that combines neuro-symbolic techniques to effectively stage proofs for Hitori puzzles, demonstrating how LLMs can guide users through complex logical sequences in an accessible manner. While our primary focus is on Hitori, the insights gained are broadly applicable to other critical decision-making domains that rely on SAT solvers for deriving and explaining complex sequences of decisions, including chemical discovery, resource allocation and scheduling, and regulatory-compliant decision-making.

\begin{figure}
  \centering
    \begin{tikzpicture}[scale=0.5,transform shape]
    \sffamily\sansmath\LARGE
    \colorlet{shaded}{blue!90!black}
    \colorlet{background}{yellow!5!white}
    \colorlet{unshaded}{green!15!white}
    \colorlet{assumption}{red}
    \begin{scope}
      \fill[background] (0,0) rectangle (5,5);
      \node[black] at (0.5,4.5) {5};
      \node[black] at (1.5,4.5) {1};
      \node[black] at (2.5,4.5) {3};
      \node[black] at (3.5,4.5) {2};
      \node[black] at (4.5,4.5) {3};
      \node[black] at (0.5,3.5) {1};
      \node[black] at (1.5,3.5) {3};
      \node[black] at (2.5,3.5) {2};
      \node[black] at (3.5,3.5) {4};
      \node[black] at (4.5,3.5) {3};
      \node[black] at (0.5,2.5) {4};
      \node[black] at (1.5,2.5) {5};
      \node[black] at (2.5,2.5) {2};
      \node[black] at (3.5,2.5) {3};
      \node[black] at (4.5,2.5) {2};
      \node[black] at (0.5,1.5) {2};
      \node[black] at (1.5,1.5) {3};
      \node[black] at (2.5,1.5) {5};
      \node[black] at (3.5,1.5) {4};
      \node[black] at (4.5,1.5) {1};
      \node[black] at (0.5,0.5) {4};
      \node[black] at (1.5,0.5) {2};
      \node[black] at (2.5,0.5) {3};
      \node[black] at (3.5,0.5) {1};
      \node[black] at (4.5,0.5) {2};
      \draw[step=1cm,ultra thin] (0,0) grid (5,5);
      \draw[thick] (0,0) rectangle (5,5);
    \end{scope}
    \begin{scope}[xshift=7cm]
      \fill[background] (0,0) rectangle (5,5);
      \fill[unshaded] (0,4) rectangle +(4,1);
      \fill[shaded] (4,4) rectangle +(1,1);
      \fill[unshaded] (0,3) rectangle +(1,1);
      \fill[shaded] (1,3) rectangle +(1,1);
      \fill[unshaded] (2,3) rectangle +(1,1);
      \fill[shaded] (3,3) rectangle +(1,1);
      \fill[unshaded] (4,3) rectangle +(1,1);
      \fill[unshaded] (0,2) rectangle +(2,1);
      \fill[shaded] (2,2) rectangle +(1,1);
      \fill[unshaded] (3,2) rectangle +(2,1);
      \fill[unshaded] (0,1) rectangle +(5,1);
      \fill[shaded] (0,0) rectangle +(1,1);
      \fill[unshaded] (1,0) rectangle +(1,1);
      \fill[shaded] (2,0) rectangle +(1,1);
      \fill[unshaded] (3,0) rectangle +(1,1);
      \fill[shaded] (4,0) rectangle +(1,1);
      \node[black] at (0.5,4.5) {5};
      \node[black] at (1.5,4.5) {1};
      \node[black] at (2.5,4.5) {3};
      \node[black] at (3.5,4.5) {2};
      \node[background] at (4.5,4.5) {3};
      \node[black] at (0.5,3.5) {1};
      \node[background] at (1.5,3.5) {3};
      \node[black] at (2.5,3.5) {2};
      \node[background] at (3.5,3.5) {4};
      \node[black] at (4.5,3.5) {3};
      \node[black] at (0.5,2.5) {4};
      \node[black] at (1.5,2.5) {5};
      \node[background] at (2.5,2.5) {2};
      \node[black] at (3.5,2.5) {3};
      \node[black] at (4.5,2.5) {2};
      \node[black] at (0.5,1.5) {2};
      \node[black] at (1.5,1.5) {3};
      \node[black] at (2.5,1.5) {5};
      \node[black] at (3.5,1.5) {4};
      \node[black] at (4.5,1.5) {1};
      \node[background] at (0.5,0.5) {4};
      \node[black] at (1.5,0.5) {2};
      \node[background] at (2.5,0.5) {3};
      \node[black] at (3.5,0.5) {1};
      \node[background] at (4.5,0.5) {2};
      \draw[step=1cm,ultra thin] (0,0) grid (5,5);
      \draw[thick] (0,0) rectangle (5,5);
    \end{scope}
  \end{tikzpicture}
  \caption{A Hitori puzzle (left) and its solution (right).}
  \vspace{-1.6em}
  \label{fig:hitori-p63}
\end{figure}

\paragraph{Why Hitori?} Hitori (see Figure~\ref{fig:hitori-p63}) is a shading puzzle played on a rectangular grid, with each cell containing a symbol, typically a number. Popularized by the renowned puzzle publisher Nikoli, Hitori has attracted significant interest not only for its recreational appeal but also for its computational aspects (it is known to be NP-complete~\cite{Hearn09}). 
The objective of the solver is to shade some of the squares to satisfy three conditions:
\begin{enumerate}[leftmargin=*,noitemsep,topsep=0pt]
\item \textbf{Uniqueness.}  No symbol appears more than once unshaded in any row or column of the grid.  For example, in Figure~\ref{fig:hitori-p63}, at least one $4$ in Column~$1$ must be shaded.
\item \textbf{Separation.}  Shaded squares cannot share an edge.  However, they may share a corner.
\item \textbf{Connection.}  The unshaded squares form an edge-connected region.  (Two squares that only touch in a corner are not edge-connected.)
\end{enumerate}

Hitori stands out as one of the simplest puzzles that naturally combines local constraints (e.g., no duplicate unshaded numbers in a row or column) with global constraints (e.g., maintaining connectivity of unshaded cells).
This combination makes Hitori an ideal candidate for exploring neuro-symbolic explanation frameworks. 
When properly posed, the proofs for local constraints can be readily translated by LLMs into natural language explanations. 
In contrast, global connectivity constraints require deeper, problem-specific insights to decompose the proof into easy-to-follow steps. 
Fortunately, Hitori's global constraints lend themselves to intuitive visualizations that reduce cognitive load during explanation, making it an excellent domain for studying how visual and symbolic reasoning interact.
Moreover, Hitori is representative of a broader class of puzzles (we have compiled a list of over 40 puzzle genres sharing these key characteristics). 
This diversity provides a rich foundation for evaluating the generalizability of our approach across problem domains.

\paragraph{What Is Proof Staging?} We formalize \emph{proof staging} as the process of organizing and sequencing explanations for the unsatisfiability of a first-order formula by assigning interpretations to constant symbols. In the spirit of puzzle solving, we focus on problems that admit a unique solution, where the goal is not just to find this solution but to explain it in a logically coherent and cognitively efficient manner.

A naive approach would first compute the solution and then explain each decision in an arbitrary order, justifying why each assignment aligns with the solver’s output. However, we argue that such arbitrary sequencing often leads to redundant explanations, where earlier steps must be repeatedly revisited to make later steps intelligible. In contrast, well-formed, humanly-solvable problem domains typically admit a natural path---or an equivalence class of paths---where explanations build incrementally on previously established facts, minimizing overlap and cognitive effort.

We posit that the proof size produced by solvers like Z3 along such a path should closely correspond to the size and clarity of natural human explanations. The key challenge, then, is how to identify this optimal explanatory path. We formalize this as the \emph{proof staging problem}, which can be viewed as finding an optimal path through the space of partial assignments of the formula. To address this, we propose heuristics for proof staging guided by insights into the \emph{weak} and \emph{strong} constraints inherent to the problem domain. These heuristics aim to produce explanations that are both logically sound and cognitively efficient, mirroring the way humans naturally build understanding through incremental reasoning.
This framework raises several critical research questions:
\begin{enumerate}[noitemsep,topsep=0pt,leftmargin=1cm,labelsep=1mm]
    \item [{\bf RQ1.}]
        Do proof staging heuristics simplify formal proofs---measured by size, reduced backtracking, structural simplicity, and ease of proof extraction---across a wide variety of Hitori puzzles?
    \item [{\bf RQ2}]
        Do the staged proofs generated by Z3 exhibit structural properties similar to human-generated explanations, particularly in terms of logical chaining of implications, minimal backtracking, and proof size? 
    \item [{\bf RQ3}]
        Can LLMs take advantage of the structured outputs produced by proof staging to generate more accurate, fluent, and helpful natural language explanations of individual solution steps?
\end{enumerate}

\paragraph{Contributions.}
We present a framework for explaining puzzle solutions, focusing on Hitori, by combining symbolic reasoning with in-context learning. 
We introduce the \emph{proof staging} problem, which structures proofs to align with human reasoning by staging subgoals to reduce complexity. To support this, we develop heuristics for identifying concise, human-understandable \emph{resolution proofs}, using natural language explanations where possible and \emph{visualizations} when reasoning becomes more complex. By classifying constraints as \emph{weak}---supporting localized reasoning---or \emph{strong}---requiring global reasoning, such as connectivity constraints in Hitori---we guide the explanation strategy accordingly. We implement this framework on top of the Z3 solver, enabling staged, parsed, and visualized proofs, and employ LLMs to produce natural language explanations that combine the rigor of symbolic solvers with the expressiveness of language models.

\section{Overview} 
\label{sec:overview}

To illustrate the potential of our approach in explaining Hitori puzzles, consider the $5{\times}5$ puzzle shown in Figure~\ref{fig:hitori-p63} along with its solution. Since this puzzle has a unique solution, asking Z3 to prove its uniqueness results in an explanations of the shading and unshading decisions. However, the corresponding resolution proof is 82,438 characters long, making it infeasible to comprehend either manually or directly through an LLM. Consequently, users of our tool begin by interactively querying the first logical step in the puzzle.

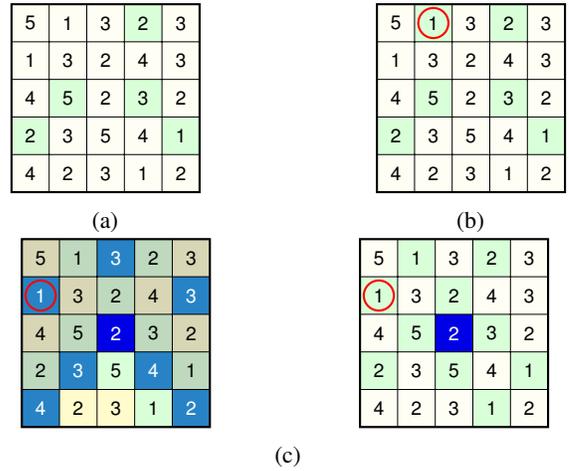
\begin{figure}[t]

     \centering
     \begin{subfigure}[b]{0.2\textwidth}
         \centering
         \begin{tikzpicture}[scale=0.5,transform shape]
    \sffamily\sansmath\LARGE
    \colorlet{shaded}{blue!90!black}
    \colorlet{background}{yellow!5!white}
    \colorlet{unshaded}{green!15!white}
    \colorlet{assumption}{red}
  
    \fill[background] (0,0) rectangle (5,5);
    \fill[unshaded] (3,4) rectangle +(1,1);
    \fill[unshaded] (1,2) rectangle +(1,1);
    \fill[unshaded] (3,2) rectangle +(1,1);
    \fill[unshaded] (0,1) rectangle +(1,1);
    \fill[unshaded] (4,1) rectangle +(1,1);
    \node[black] at (0.5,4.5) {5};
    \node[black] at (1.5,4.5) {1};
    \node[black] at (2.5,4.5) {3};
    \node[black] at (3.5,4.5) {2};
    \node[black] at (4.5,4.5) {3};
    \node[black] at (0.5,3.5) {1};
    \node[black] at (1.5,3.5) {3};
    \node[black] at (2.5,3.5) {2};
    \node[black] at (3.5,3.5) {4};
    \node[black] at (4.5,3.5) {3};
    \node[black] at (0.5,2.5) {4};
    \node[black] at (1.5,2.5) {5};
    \node[black] at (2.5,2.5) {2};
    \node[black] at (3.5,2.5) {3};
    \node[black] at (4.5,2.5) {2};
    \node[black] at (0.5,1.5) {2};
    \node[black] at (1.5,1.5) {3};
    \node[black] at (2.5,1.5) {5};
    \node[black] at (3.5,1.5) {4};
    \node[black] at (4.5,1.5) {1};
    \node[black] at (0.5,0.5) {4};
    \node[black] at (1.5,0.5) {2};
    \node[black] at (2.5,0.5) {3};
    \node[black] at (3.5,0.5) {1};
    \node[black] at (4.5,0.5) {2};
    \draw[step=1cm,ultra thin] (0,0) grid (5,5);
    \draw[thick] (0,0) rectangle (5,5);
  \end{tikzpicture}
         \caption{}
         \label{fig:sandwich}
     \end{subfigure}
     \hfill
     \begin{subfigure}[b]{0.2\textwidth}
         \centering
  \begin{tikzpicture}[scale=0.5,transform shape]
    \sffamily\sansmath\LARGE
    \colorlet{shaded}{blue!90!black}
    \colorlet{background}{yellow!5!white}
    \colorlet{unshaded}{green!15!white}
    \colorlet{assumption}{red}
    \colorlet{latest}{red}
    \fill[background] (0,0) rectangle (5,5);
    \fill[unshaded] (3,4) rectangle +(1,1);
    \fill[unshaded] (1,2) rectangle +(1,1);
    \fill[unshaded] (3,2) rectangle +(1,1);
    \fill[unshaded] (0,1) rectangle +(1,1);
    \fill[unshaded] (4,1) rectangle +(1,1);
    \fill[unshaded] (1,4) rectangle +(1,1);
    \node[black] at (0.5,4.5) {5};
    \node[black] at (1.5,4.5) {1};
    \node[black] at (2.5,4.5) {3};
    \node[black] at (3.5,4.5) {2};
    \node[black] at (4.5,4.5) {3};
    \node[black] at (0.5,3.5) {1};
    \node[black] at (1.5,3.5) {3};
    \node[black] at (2.5,3.5) {2};
    \node[black] at (3.5,3.5) {4};
    \node[black] at (4.5,3.5) {3};
    \node[black] at (0.5,2.5) {4};
    \node[black] at (1.5,2.5) {5};
    \node[black] at (2.5,2.5) {2};
    \node[black] at (3.5,2.5) {3};
    \node[black] at (4.5,2.5) {2};
    \node[black] at (0.5,1.5) {2};
    \node[black] at (1.5,1.5) {3};
    \node[black] at (2.5,1.5) {5};
    \node[black] at (3.5,1.5) {4};
    \node[black] at (4.5,1.5) {1};
    \node[black] at (0.5,0.5) {4};
    \node[black] at (1.5,0.5) {2};
    \node[black] at (2.5,0.5) {3};
    \node[black] at (3.5,0.5) {1};
    \node[black] at (4.5,0.5) {2};
    \draw[latest,thick] (1.5,4.5) circle[radius=0.4cm];
    \draw[step=1cm,ultra thin] (0,0) grid (5,5);
    \draw[thick] (0,0) rectangle (5,5);
  \end{tikzpicture}
         \caption{}
         \label{fig:complex-rule}
     \end{subfigure}
     \hfill
     \begin{subfigure}[b]{0.4\textwidth}
         \centering
    \begin{tikzpicture}[scale=0.5,transform shape]
  \sffamily\sansmath\LARGE
  \colorlet{shaded}{blue!90!black}
  \colorlet{background}{yellow!5!white}
  \colorlet{done}{green!15!white}
  \colorlet{reached}{gray}
  \colorlet{trueassumption}{yellow!25!white}
  \colorlet{falseassumption}{RoyalBlue!80!white}
  \colorlet{latest}{red}
  \fill[background] (0,0) rectangle (5,5);
  \fill[trueassumption] (0,4) rectangle +(1,1);
  \fill[reached,opacity=0.3] (0,4) rectangle +(1,1);
  \node[black] at (0.5,4.5) {5};
  \fill[done] (1,4) rectangle +(1,1);
  \fill[reached,opacity=0.3] (1,4) rectangle +(1,1);
  \node[black] at (1.5,4.5) {1};
  \fill[falseassumption] (2,4) rectangle +(1,1);
  \node[white] at (2.5,4.5) {3};
  \fill[done] (3,4) rectangle +(1,1);
  \fill[reached,opacity=0.3] (3,4) rectangle +(1,1);
  \node[black] at (3.5,4.5) {2};
  \fill[trueassumption] (4,4) rectangle +(1,1);
  \fill[reached,opacity=0.3] (4,4) rectangle +(1,1);
  \node[black] at (4.5,4.5) {3};
  \fill[falseassumption] (0,3) rectangle +(1,1);
  \node[white] at (0.5,3.5) {1};
  \fill[trueassumption] (1,3) rectangle +(1,1);
  \fill[reached,opacity=0.3] (1,3) rectangle +(1,1);
  \node[black] at (1.5,3.5) {3};
  \fill[done] (2,3) rectangle +(1,1);
  \fill[reached,opacity=0.3] (2,3) rectangle +(1,1);
  \node[black] at (2.5,3.5) {2};
  \fill[trueassumption] (3,3) rectangle +(1,1);
  \fill[reached,opacity=0.3] (3,3) rectangle +(1,1);
  \node[black] at (3.5,3.5) {4};
  \fill[falseassumption] (4,3) rectangle +(1,1);
  \node[white] at (4.5,3.5) {3};
  \fill[trueassumption] (0,2) rectangle +(1,1);
  \fill[reached,opacity=0.3] (0,2) rectangle +(1,1);
  \node[black] at (0.5,2.5) {4};
  \fill[done] (1,2) rectangle +(1,1);
  \fill[reached,opacity=0.3] (1,2) rectangle +(1,1);
  \node[black] at (1.5,2.5) {5};
  \fill[shaded] (2,2) rectangle +(1,1);
  \node[background] at (2.5,2.5) {2};
  \fill[done] (3,2) rectangle +(1,1);
  \fill[reached,opacity=0.3] (3,2) rectangle +(1,1);
  \node[black] at (3.5,2.5) {3};
  \fill[trueassumption] (4,2) rectangle +(1,1);
  \fill[reached,opacity=0.3] (4,2) rectangle +(1,1);
  \node[black] at (4.5,2.5) {2};
  \fill[done] (0,1) rectangle +(1,1);
  \fill[reached,opacity=0.3] (0,1) rectangle +(1,1);
  \node[black] at (0.5,1.5) {2};
  \fill[falseassumption] (1,1) rectangle +(1,1);
  \node[white] at (1.5,1.5) {3};
  \fill[done] (2,1) rectangle +(1,1);
  \node[black] at (2.5,1.5) {5};
  \fill[falseassumption] (3,1) rectangle +(1,1);
  \node[white] at (3.5,1.5) {4};
  \fill[done] (4,1) rectangle +(1,1);
  \fill[reached,opacity=0.3] (4,1) rectangle +(1,1);
  \node[black] at (4.5,1.5) {1};
  \fill[falseassumption] (0,0) rectangle +(1,1);
  \node[white] at (0.5,0.5) {4};
  \fill[trueassumption] (1,0) rectangle +(1,1);
  \node[black] at (1.5,0.5) {2};
  \fill[trueassumption] (2,0) rectangle +(1,1);
  \node[black] at (2.5,0.5) {3};
  \fill[done] (3,0) rectangle +(1,1);
  \node[black] at (3.5,0.5) {1};
  \fill[falseassumption] (4,0) rectangle +(1,1);
  \node[white] at (4.5,0.5) {2};
  \draw[latest,thick] (0.5,3.5) circle[radius=0.4cm];
  \draw[step=1cm,ultra thin] (0,0) grid (5,5);
  \draw[thick] (0,0) rectangle (5,5);
\end{tikzpicture}
\hfill
\begin{tikzpicture}[scale=0.5,transform shape]
  \sffamily\sansmath\LARGE
  \colorlet{shaded}{blue!90!black}
  \colorlet{background}{yellow!5!white}
  \colorlet{done}{green!15!white}
  \colorlet{reached}{gray!15!white}
  \colorlet{trueassumption}{yellow!25!white}
  \colorlet{falseassumption}{RoyalBlue!80!white}
  \colorlet{latest}{red}
  \fill[background] (0,0) rectangle (5,5);
  \node[black] at (0.5,4.5) {5};
  \fill[done] (1,4) rectangle +(1,1);
  \node[black] at (1.5,4.5) {1};
  \node[black] at (2.5,4.5) {3};
  \fill[done] (3,4) rectangle +(1,1);
  \node[black] at (3.5,4.5) {2};
  \node[black] at (4.5,4.5) {3};
  \fill[done] (0,3) rectangle +(1,1);
  \node[black] at (0.5,3.5) {1};
  \node[black] at (1.5,3.5) {3};
  \fill[done] (2,3) rectangle +(1,1);
  \node[black] at (2.5,3.5) {2};
  \node[black] at (3.5,3.5) {4};
  \node[black] at (4.5,3.5) {3};
  \node[black] at (0.5,2.5) {4};
  \fill[done] (1,2) rectangle +(1,1);
  \node[black] at (1.5,2.5) {5};
  \fill[shaded] (2,2) rectangle +(1,1);
  \node[background] at (2.5,2.5) {2};
  \fill[done] (3,2) rectangle +(1,1);
  \node[black] at (3.5,2.5) {3};
  \node[black] at (4.5,2.5) {2};
  \fill[done] (0,1) rectangle +(1,1);
  \node[black] at (0.5,1.5) {2};
  \node[black] at (1.5,1.5) {3};
  \fill[done] (2,1) rectangle +(1,1);
  \node[black] at (2.5,1.5) {5};
  \node[black] at (3.5,1.5) {4};
  \fill[done] (4,1) rectangle +(1,1);
  \node[black] at (4.5,1.5) {1};
  \node[black] at (0.5,0.5) {4};
  \node[black] at (1.5,0.5) {2};
  \node[black] at (2.5,0.5) {3};
  \fill[done] (3,0) rectangle +(1,1);
  \node[black] at (3.5,0.5) {1};
  \node[black] at (4.5,0.5) {2};
  \draw[latest,thick] (0.5,3.5) circle[radius=0.4cm];
  \draw[step=1cm,ultra thin] (0,0) grid (5,5);
  \draw[thick] (0,0) rectangle (5,5);
\end{tikzpicture}
         \caption{}
         \label{fig-connectivity}
     \end{subfigure}
        \caption{Stages in solving the Hitori: (a) sandwich rule for cells $\texttt{r1c4}$ (and others) shaded in light green; (b) multi-step deduction for \texttt{r1c2} to be unshaded; and (3) A picture based connectivity proof for the cell $\texttt{r2c1}$ to remain unshaded.}
         \vspace{-1em}
        \label{fig:hitori-graphs}
\end{figure}
  
A commonly useful pattern in a Hitori grid is the \emph{sandwich}, where a symbol is surrounded---either horizontally or vertically---by two identical symbols. 
One of these two identical symbols must be shaded, which forces the sandwiched symbol to remain unshaded due to adjacency. 
In our example, there are five sandwiched squares.  Notably, \texttt{r3c4} (the grid square in Row~$3$, Column~$4$) is sandwiched both horizontally and vertically.
The sample proof provided by the tool for cell \texttt{r1c4} is given in the following box.

\begin{mybox}{\texttt{r1c4}}
\footnotesize{\textbf{Z3 Proof} for \texttt{r1c4} unshaded:}
 \begin{lstlisting}[basicstyle=\scriptsize\ttfamily]
c01_04
\end{lstlisting}
\vspace{0.5em}
\footnotesize{\textbf{LLM Explanation:} This cell lies between two identical numbers (3s) in its row. According to the Sandwich Lemma, if two identical numbers are separated by exactly one cell, that middle cell must be unshaded. Shading it would allow both 3s on either side to remain unshaded, violating the Uniqueness Rule. Therefore, (1,4) must be unshaded.}

\end{mybox}
  
From its prompt, the LLM knows that the unit clause \texttt{c01\_04} encodes an application of the sandwich rule.  The interaction can continue as the user selects additional cells or requests the next step from the tool. Eventually, this leads to more complex deductions, such as the following:
\begin{mybox}{\texttt{r1c2}}
\footnotesize{\textbf{Z3 Proof} for \texttt{r1c2} unshaded:}
    \begin{lstlisting}[basicstyle=\scriptsize\ttfamily]
(let ((a!1 (unit-resolution (asserted (or (not c02_02) (not c02_05)))
          (unit-resolution (asserted (or c01_02 c02_02))
                          (asserted (not c01_02))
                          c02_02)
          (not c02_05)))
      (a!2 (unit-resolution (asserted (or (not c01_03) (not c01_05)))
          (unit-resolution (asserted (or c01_02 c01_03))
                          (asserted (not c01_02))
                          c01_03)
          (not c01_05))))
  (unit-resolution (asserted (or c01_05 c02_05)) a!1 a!2 false))
\end{lstlisting}
\vspace{0.5em}
\footnotesize{\textbf{LLM Explanation:}  
This cell must be \emph{unshaded}. If \texttt{r1c2} were shaded, then:
\begin{itemize}
    \item \texttt{r1c3} would have to be unshaded to satisfy the Separation Rule.
    \item But \texttt{r1c3} and \texttt{r1c5} have the same symbol (3), so both cannot remain unshaded. This forces \texttt{r1c5} to be shaded.
    \item Similarly, shading \texttt{r1c2} forces \texttt{r2c2} to be unshaded due to adjacency.
    \item Now, \texttt{r2c2} and \texttt{r2c5} have the same symbol (3), so one must be shaded. Since \texttt{r1c5} is already shaded, \texttt{r2c5} must remain unshaded.
    \item This leads to \texttt{r1c5} and \texttt{r2c5} both being shaded and adjacent, violating the Separation Rule.
\end{itemize}}

Thus, assuming \texttt{r1c2} is shaded leads to a contradiction. Therefore, it must be unshaded.

\end{mybox}

The next proof concerns cell \texttt{r2c1}, which must remain unshaded. The reason for this is that assuming \texttt{r2c1} to be shaded leads to a violation of the connectivity constraint. This is illustrated in Figure~\ref{fig:hitori-graphs}(c):
The left subfigure shows that if \texttt{r2c1} is shaded, the following cells must also be shaded by logical implication: \texttt{r5c1}, \texttt{r4c2}, \texttt{r4c4}, and \texttt{r5c5}.
Combined with cell \texttt{r3c3}, which is already known to be shaded, this results in a disconnected grid, violating Hitori's connectivity rule.
The right subfigure in Figure~\ref{fig:hitori-graphs}(c) depicts that, to preserve connectivity, cell \texttt{r2c1} must remain unshaded.

\section{Preliminaries}
\label{sec:prelims}







A \emph{propositional formula} is built recursively from a set of \emph{propositional variables} $V$ by applying Boolean connectives.  We take as Boolean connectives $\neg$ (negation), $\wedge$ (conjunction) and $\vee$ (disjunction) in order of decreasing precedence, with their usual semantics.  A \emph{literal} is either a variable or its negation.  A \emph{clause} is the disjunction of a set of literals.  A formula in \emph{conjunctive normal form} (CNF) is the conjunction of a set of clauses.   A formula is \emph{satisfiable} if there is an assignment of $\top$ (true) and $\bot$ (false) to the variables appearing in it that makes the formula evaluate to $\top$.  Such an assignment is a \emph{model} of the formula.  A formula is \emph{valid} (or a \emph{tautology}) if it evaluates to $\top$ for all assignments to its variables.  A formula is valid if its negation is not satisfiable.  

SAT solvers often play a central role in propositional reasoning.  In what follows, we focus on CNF formulae.  By SAT solvers, we mean CNF SAT solvers; specifically those based on the CDCL (Conflict-Driven Clause-Learning) algorithm \cite{Silva99c}.
SAT solvers, in addition to answering the satisfiability question, are often required to provide a certificate for their decision.  Certificates, besides allowing independent verification of a solver's conclusion, have numerous applications.  For a satisfiable formula, a model is a certificate.  Evidence of unsatisfiability, however, is usually provided in one of two forms: 
\begin{itemize}[leftmargin=*,noitemsep,topsep=0pt]
    \item A \emph{resolution proof}, in which a contradiction (the empty clause) is derived from the formula's clauses by repeated application of the resolution rule of inference, to be discussed presently.
    \item An \emph{unsatisfiable core}, which is a subset of the clauses that is, by itself, unsatisfiable.  A common special case occurs when the clauses in the core are literals from a set of \emph{assumptions}.
\end{itemize}
The preferred type of certificate depends on the intended use.  For our application, resolution proofs are the natural choice.  Given two clauses, $\gamma_1 = C_1 \vee v$ and $\gamma_2 = C_2 \vee \neg v$ such that $v$ and $\neg v$ do not appear in $C_1$ and $C_2$, the \emph{resolvent} of $\gamma_1$ and $\gamma_2$ is the clause obtained from $C_1 \vee C_2$ by removing duplicate literals.  Since the resolvent of two clauses is implied by their conjunction, the resolution rule is sound.  Resolution is also \emph{refutationally complete} \cite{Davis60,Robins65}:  The empty clause may be obtained if, and only if, the conjunction of the clauses is unsatisfiable.

A resolution in which at least one of the resolved clauses is a literal, is a \emph{unit} resolution.  A proof by unit resolution is one in which all inferences are unit resolutions.  A proof by \emph{linear} resolution is one in which one of the resolved clauses is the most recently computed resolvent (except for the first step of the proof).  Linear resolution is complete. Unit resolution is desirable because the resolvent is simpler than at least one input clause.  Unfortunately, it is not complete.  

SAT solvers use unit resolution to effect \emph{Boolean constraint propagation}.  They supplement it with backtracking search and conflict analysis to achieve a complete proof system.  Conflict analysis consists of repeated resolutions, so that, when the CNF formula is unsatisfiable, from these resolution steps a proof (or an unsatisfiable core) can be assembled.  If no branching occurred during the search and no model is found, the proof of unsatisfiability is by unit resolution.  Related to the ability to produce a certificate for an unsatisfiable formula is the ability to compute the \emph{backbone} literals of a satisfiable formula \cite{Kilby05}, that is, those literals that are true in all the formula's models.  Indeed, efficient algorithms for their computation rely on unsatisfiable cores \cite{Janota15}.

Since resolution proofs produced by SAT solvers may be quite large \cite{Heule16}, efficient formats have been devised for their representation \cite{Heule15,Kiesl18}.  Those formats, however, are optimized for mechanical proof checking.  Therefore, in this work, we prefer the less concise but more readable proofs produced by Z3 \cite{Moura11} and described in \cite{Moura08b}.  Z3 combines a CDCL engine and an array of \emph{theory solvers} into a Satisfiability Modulo Theories (SMT) solver, which can be applied to first-order formulae involving arithmetic and data structures.  In this work, we only use the propositional capabilities of Z3.  Hence, we could use any proof-producing SAT solver that can compute backbone literals, like CaDiCaL \cite{Biere24}.





\section{Proof Staging for Hitori}\label{sec:proof-staging}



In a proof-staging problem, we are given a propositional formula with a unique model and aim to prove the value of each variable in that model. We are also given a set of subgoals. For Hitori, a subgoal may be to prove whether a grid square is shaded or unshaded; for Sudoku, it might involve proving that a square contains a specific value or that a candidate can be eliminated. Achieving all subgoals must entail a complete explanation of the model.

Given our formula and the associated set of subgoals, proof staging is the process of ordering the proof subgoals and choosing a proof technique for each subgoal so as to get a simple proof.  The choice of the proof technique is dynamic: it happens when the subgoal is attempted.  This contributes to the difficulty of picking a good order.  For Hitori, we rely on two proof techniques: propositional resolution proofs and connectivity proofs.  For classic Sudoku, resolution is enough, but for Sudoku variants, which, with few exceptions, are not simple coloring problems, we may need arithmetic.


\subsection{Encoding Hitori}
\label{sec:encoding-for-sat}  

From a Hitori grid, we obtain a propositional formula that is satisfiable if, and only if, the puzzle has a solution.  We briefly summarize the process here, focusing on the aspects that are relevant to the explanation of a puzzle.  More details may be found in Appendix~\ref{app:hitori-encoding}.


We associate a Boolean variable, $c_{i,j}$, to cell (or \emph{square}) $(i,j)$ of an $m \times n$ Hitori grid , stipulating that $c_{i,j}$ is true if, and only if, the cell is clear (unshaded).  If $g_{i,j}$ is the symbol in cell $(i,j)$, then the uniqueness rule is encoded by the two-literal clauses $\neg c_{i,j} \vee \neg c_{k,\ell}$ for all $i,j,k,\ell$ such that $(i,j)$ and $(k,\ell)$ are distinct cells in the same row or the same column, and $g_{i,j} = g_{k,\ell}$.  The separation rule is encoded by the two-literal clauses $c_{i,j} \vee c_{k,\ell}$ for all $i,j,k,\ell$ such that $(i,j)$ and $(k,\ell)$ are distinct cells that share an edge.

Encoding the connection rule for SAT can be done in several ways, none of which is entirely straightforward.  The approach we follow lets the SAT solver construct a spanning tree of the unshaded cell graph.  To that effect, we associate a \emph{parent pointer} to each cell.  Each pointer is encoded by four Boolean variables.  A fifth variable is used to prevent the parent pointers from forming cycles, according to the technique of \cite{Brock18}.  The SAT solver also chooses an unshaded cell as the root of the tree.  
Full details of the encoding, which are given in the Appendix, are not important for the explanation of the puzzles, because the clauses for the connection rule do not appear in them as discussed in Section~\ref{sec:lowest-hanging}.

Since a connected graph may have many spanning trees, the map from constraint models to puzzle solutions is not injective.  However, to check uniqueness of the puzzle solution given a model of the constraints, it suffices to add to the constraints a \emph{blocking clause}.  This clause asserts that a model should differ from the one that was found in the values of one or more $c_{i,j}$ variables.


To the clauses described so far, we add the following redundant ones to simplify the explanations. These are derived from straightforward consequences of Hitori’s rules.

\paragraph{Sandwich lemma.}  A square \emph{sandwiched} between two squares that carry the same symbol---either in the same row or in the same column---must remain unshaded.  This is because at least one of the neighbors must be shaded by the uniqueness rule; hence, the sandwiched cell must remain unshaded by the separation rule.  This lemma produces unit clause $c_{i,j}$ for each sandwiched cell $(i,j)$.

\paragraph{Unshaded Neighbor lemma.} If $mn \geq 4$, every square in the Hitori grid must have a neighbor that is not shaded.  If the square is shaded, this follows from the separation rule.  If square $(i,j)$ is unshaded and $mn \geq 4$, then there is at least another unshaded square in the grid that must connect to $(i,j)$.  
This is impossible unless $(i,j)$ has a neighbor that is not shaded.  The existence of an unshaded neighbor is easily encoded by a clause with a positive literal for each neighbor of $(i,j)$.  While this lemma only provides a necessary condition for connectedness, it takes care of a large fraction of the proofs that involve the connection rule.

Other lemmas could be used to capture known solution techniques.  We only discuss one.
Let a \emph{seed} of a Hitori puzzle be a square whose symbol does not appear elsewhere in the square's row and column.  \citet[p.\ 182]{Knuth-v4b} observes that if a puzzle has a unique solution, then all seeds are unshaded in that solution and all seeds that are not adjacent to shaded squares are articulation points for the graph of the unshaded squares.  However, no further deductions are enabled from the unshading of the seeds.  Moreover, our objective is not to explain how to find \emph{some} solution, but to explain how to find \emph{the} solution.  Hence, we do not rely on uniqueness assumptions that would have to be later discharged.  Accordingly, we do not make use of Knuth's observations.


\subsection{A Greedy Heuristic for Proof Staging}
\label{sec:lowest-hanging}

In Section~\ref{sec:encoding-for-sat}, we have shown how a Hitori puzzle may be encoded into a propositional formula $f$ that is satisfiable if, and only if, the puzzle has a solution.  We have further shown how uniqueness of the solution may be checked by conjoining a blocking clause $b$ to $f$ and checking whether the resulting formula is unsatisfiable.  If $f \wedge b$ is indeed unsatisfiable, a resolution proof of that outcome, which we call a \emph{monolithic} proof, provides an explanation for the solution of the puzzle, albeit one that is seldom adequate.  The last column of Table~\ref{tab:some-hard-puzzles} shows that the monolithic proof is usually so large as to overwhelm both humans and LLMs.  

One could think of decomposing the proof into lemmas, one for each cell of the grid.
Indeed, for each literal $\ell$ that appears in the blocking clause $b$, the monolithic proof contains a refutation that $\ell$ can be made true while satisfying $f$.  Proof decomposition, however, is ineffective for two reasons.  The first reason is that the blocking clause, due to its $mn$ literals, is a weak constraint: one that does not immediately cause the SAT solver to deduce values.  In contrast, asking whether a specific $c_{i,j}$ variable may have its value reversed with respect to the (unique) puzzle solution adds to $f$ a unit literal (a clause with one literal) which often triggers deductions.  The net result is that the monolithic proof often contains unnecessary case splitting, which makes it hard to follow.

The second reason is even more important.  Most inferences in the solution of a Hitori puzzle only rely on the uniqueness and separation rules.   Those inferences are usually explained concisely by resolution proofs.  The same applies to those (local) inferences that can be justified by the Unshaded Neighbor lemma.  The inferences that depend on the connection rule, but cannot be explained by the Unshaded Neighbor lemma, however, often require very unwieldy resolution proofs.  On the one hand, these proofs involve all the variables used to encode the problem.  Specifically, they involve the constraints on the parent pointers.  On the other hand, these proofs must show that, no matter where the parent pointers point, a contradiction is reached.  This often leads to extensive case splitting, which results in hard-to-read proofs.  In summary,
\begin{itemize}[leftmargin=*,topsep=0pt,noitemsep]
    \item Monolithic proofs do not provide a good platform for the explanation of Hitori puzzles.
    \item Propositional resolution proof are ill suited to explain deductions prompted by the connection rule.
\end{itemize}
We address both problems via proof staging.  As mentioned earlier, the propositional formula is $f$, and the subgoals are the shading values of all cells in the grid.  The order in which the subgoals are tackled is greedy: subgoals that only depend on the uniqueness and separation rules are given priority over subgoals that depend on the connection rule.  Ties among the higher priority subgoals are broken by the size of their respective resolution proofs.   For the low priority subgoals, we analyze the connectivity of the grid graph and we present the results graphically, instead of textually.  We call these graphical forms \emph{proofs by picture}.


\paragraph{Strong and Weak Solutions.} A straightforward but inefficient way to identify the lowest hanging fruit among the subgoals is to generate all their proofs and pick the subgoal with the shortest one.  Once a subgoal has been explained, it becomes a unit literal.  This simplifies the proofs for other subgoals.  Hence, in the next round, all remaining subgoals are ranked anew.  Instead, we rely on the notion of weak and strong constraints.  

A weak solution for a Hitori puzzle is an assignment to the $c_{i,j}$ variables that complies with the uniqueness and separation rules, but may violate the connection rule.  By analogy, a \emph{weak constraint}
is one that only uses the $c_{i,j}$ variables, and none of the remaining variables.  The uniqueness and separation constraints are fully expressible in terms of the $c_{i,j}$ variables.  Among the connection
constraints, only those that rely on the Unshaded Neighbor lemma are so expressible.  In contrast, a strong constraint is one that depends on other variables besides the $c_{i,j}$ variables.  A \emph{weak solver} is a solver that is only provided the weak constraint of the puzzle, while a \emph{strong solver} has access to all constraints, both weak and strong.  A proof produced by a weak solver is a \emph{weak proof}.

The solution to the puzzle is computed by a strong SAT solver, while the selection of the next subgoal to tackle makes use of a weak solver.   Goals whose proofs require strong constraints result in satisfiable sets of weak constraints.  Moreover, the weak constraints are only a fraction of all constraints and are much easier to solve.
Indeed, unit resolution is complete for the weak constraints (if we ignore the \emph{Unshaded Neighbor} clauses) because they only require two-literal clauses.  Unit resolution is not complete, in general, for 2CNF.  The classic example is 
\[
(a \vee b) \wedge (a \vee \neg b) \wedge (\neg a \vee b) \wedge (\neg a \vee \neg b),
\]
which has no unit-resolution refutation.  However, in our proofs, we always have at least one unit literal, given by the negation of the subgoal.  We can then rely on the completeness of \emph{linear resolution} \cite{Anders70}.  Linear resolution, starting with a unit literal, produces resolvents that are all unit literals because the side clauses have no more than two literals.  Therefore, all steps are unit resolutions.
Even with the inclusion of the clause based on the Unshaded Neighbor lemma, we empirically observe that weak proofs only need unit resolutions.

\paragraph{Proofs by Picture.} In the solution of a Hitori puzzle, the connection rule usually plays a secondary role, or no role at all, until enough cells have been shaded.  
That is why taking up subgoals that depend on that rule after the subgoals with weak proofs usually works.  Explaining a subgoal by a connectivity argument often enables more weak proofs, so that in the final stages of a proof, weak proofs and proofs by picture alternate.  If a subgoal has no weak proof, it means that the weak constraints, conjoined with the negation of the subgoal, are satisfiable.  We ask the weak solver for the backbone literals of those satisfiable constraints.  These consequences of the negation of the goal may lead to a disconnected grid graph, which proves that the goal value cannot be flipped, or to a graph that has articulation points \cite{Hopcro73}.  In the latter case, the constraint that the articulation points must be left unshaded is added to the weak solver as additional consequences of the negation of the subgoal.  If the resulting constraints become unsatisfiable, the subgoal is proved.  If there are no articulation points or the constraints remain satisfiable, we need to resort to a strong proof.  Empirically, this has never happened in our experiments.  We conjecture that this is because puzzles that would require those strong proofs are unsuitable for human consumption.


\section{Experimental Results}
\label{sec:results}



\begin{table}
\small
    \centering
    \caption{Puzzles stats. For each puzzle, \emph{size} gives the size of the grid; \emph{avg} the arithmetic mean of the lengths of the resolution proofs; \emph{max} the length of the longest resolution proof; \emph{pbp} the number of proofs by picture; and \emph{uniq} the length of a (monolithic) resolution proof that the solution is unique.  Proof lengths are in characters.}
    \begin{tabular}{lccccc}
    \toprule
        \textbf{puzzle} & \textbf{size} & \textbf{avg }& \textbf{max} & \textbf{pbp} & \textbf{uniq} \\\hline
        p51 & $4 \times 4$ &  134.67 & 223 & 1 & \num{4511} \\
        p63 & $5 \times 5$ & 159.2  & 518 & 5 & \num{82438} \\
        p107 & $5 \times 5$  & 202.21 & 777 & 1 & \num{7442} \\
        p43  & $8 \times 4$ & 174.84 & 518 & 0 & \num{3834} \\
        p48  & $11 \times 3$  & 156.14 & 664 & 4 & \num{157507} \\
        p9  & $8 \times 8$ & 114.69 & 416 & 3 & \num{48298} \\
        p111 & $8 \times 8$ & 113.72 & 471 & 4 & \num{232362} \\
        p29  & $8 \times 8$ & 140.48 & 640 & 8 & \num{72913} \\
        p34  & $10 \times 10$ & 119.19 & 601 & 4 & \num{549888} \\
        p65  & $10 \times 10$ & 164.06 & 549 & 4 & \num{874386} \\
        p105 & $10 \times 10$ & 122.29 & 549 & 4 & \num{28086} \\
        p36  & $15 \times 15$ & 104.69 & 601 & 5 & \num{212337} \\
        p25  & $25 \times 25$ & 102.14 & 188 & 7 & \num{4091665} \\
        \bottomrule
    \end{tabular}
    \label{tab:some-hard-puzzles}
\end{table}

\subsection{Proof Staging (RQ1 and RQ2)}
We assembled a data set of $107$ Hitori puzzles from various sources; of these, $15$ were mechanically generated by us.  All have unique solutions.
Table~\ref{tab:some-hard-puzzles} presents statistics for the puzzles that we found more challenging, due to the complexity of their resolution proofs (high \emph{avg} and \emph{max}) or the difficulty of proving uniqueness (large \emph{uniq} values). High \emph{pbp} counts further indicate cases where visual reasoning is needed, highlighting the limits of purely symbolic explanations.  

As an example, consider \texttt{p25}, which is challenging for humans due to its sheer size.  The low values of \emph{avg} and \emph{max}, however, means that, while scanning the grid may be taxing, the logical steps are all simple.  Out of $625$ subgoals, only $7$ require proofs by picture.  This means that the connection rule does not play a major role in the solution.  Even such a small number of subgoals related to connectivity, though, is enough to blow up the monolithic proof to the point of making it useless for explanation.  Note that the total length of the $618$ resolution proofs resulting from the staging approach is $63,121$ characters.  While the contrast between monolithic proof and staged proof steps is exacerbated by the grid size, it is also notable in small puzzles that require more intricate reasoning, like \texttt{p63}, where the combined length of all proof steps is more that $25$ times shorter than the monolithic proof.

These puzzles provide a rigorous test for {\bf RQ1} and {\bf RQ2}. Despite their complexity, proof staging heuristics reduce proof size and backtracking, simplifying extraction and explanation. The staged proofs also exhibit structural patterns such as incremental reasoning, logical chaining, and visual aids that align with human explanatory strategies.



\subsection{Explanation Generation (RQ3)} 
We designed a small study to test the ability of LLMs to generate accurate, fluent, and helpful natural language explanations of individual solutions steps when provided with the structured output produced by the proof staging mechanism. In addition, we evaluated the impact of the length and complexity of the proof steps on generation performance.


\paragraph{Model and Prompting Strategy.} We use \texttt{DeepSeek R1}, an open-source LLM with strong performance on code-language tasks~\cite{deepseekai2025deepseekr1incentivizingreasoningcapability}. Using a few-shot prompting approach~\cite{NEURIPS2020_1457c0d6}, we provide the model with three demonstrations of single-step proof objects paired with natural language explanations. To interpret proof objects, the LLM must understand the puzzle constraints and reason about the board state—tracking shaded and unshaded cells and how they inform each explanation. To support this, we augment the prompt with two previous solution steps, the current board state (i.e., shaded/unshaded cells), and the fixed set of clauses and definitions used in the SAT encoding. An example prompt is provided in the Appendix.

\paragraph{Data.} We sampled $5$ proof steps and their explanations from the $55$ Hitori puzzles, resulting in $275$ explanation steps. These steps occur at different stages of the proof with varying lengths and levels of complexity. In general, steps that occur in the earlier stages of a proof are more trivial than the ones that occur in the later stages. To account for this, we split our data into two subsets. The first set contains trivial steps that typically occur at the start of the proof, while the second set contains challenging steps that occur towards the end of the proof. To further study the impact of structured context, we experiment with two configurations, one in which the model sees the entire history of shaded and unshaded cells up to that step and one in which we only provide the history of the cells that are relevant to the individual proof step. 

\paragraph{Evaluation Metrics.}
In order to assess the quality of the explanations, we used expert ratings. Each step was rated by two annotators. Table~\ref{tab:translation-eval} outlines the evaluation criteria, which were rated using a Likert scale from 1 to 5.
\begin{table}[t]
\small
    \centering
    \caption{Evaluation Criteria for Explanation Quality}
    \vspace{0.5em}
    \resizebox{\columnwidth}{!}{%
    \begin{tabular}{lp{6cm}}
    \toprule
        \textbf{Criterion} & \textbf{Description} \\\hline
        A. Correctness & The explanation is correct. \\
        \midrule
        B. Relevance & The explanation \textit{only} uses facts necessary to understand the step. \\
        \midrule
        C. Completeness & The explanation addresses \textit{all} clauses in the proof needed for a good explanation. \\
        \midrule
        D. Falsities & The explanation contains assertions that are simply not true. \\
        \midrule
        E. Clarity & The explanation is lucid and helps the reader understand the proof. \\
        \midrule
        F. Conciseness & The explanation is reasonably short, avoiding unnecessary repetitions. \\
        \bottomrule
    \end{tabular}}
    \label{tab:translation-eval}
\end{table}






\paragraph{Results for Trivial Steps.} We observe a relatively low average correctness rating for trivial steps ($A = 2.96-3.34-$, where $2$ means disagree and $3$ means neither agree nor disagree). Interestingly, we observed that the annotators agreed less when assessing the proof steps that were trivial, obtaining a quadratically weighted Cohen's Kappa of $0.42$, suggesting moderate agreement. Upon analysis, we observe that the language produced by the LLM to explain succinct and trivial steps is often imprecise. The disagreement stemmed from differences in leniency towards this impreciseness. This had knock-on effects on subsequent criteria, as one annotator assumed all statements in incorrect proofs to be unhelpful to the necessary conclusion. For example, we observe that relevance scores drop after an evaluator has judged a translation incorrect because any evidence presented by the LLM is deemed to be ``irrelevant''. The succinctness of the proof, as well as the lack of previous states, results in the LLM ``overthinking'', often generating a non-factual explanation in the process. 
We also observe that performance slightly improves when only states pertinent to the step are provided to the LLM, possibly due to reduced context burdens.

\paragraph{Results for Complex Steps.} We observe good average scores in most metrics for complex steps (above $4$ for positive criteria and below $2$ for negative criteria, where $4$ means disagree and $2$ means disagree), with the exception of the conciseness metric ($F = 3.12-3.28$). We also observed greater agreement between the evaluators, achieving Cohen's Kappa of $0.6$. This is due in part to the fact that later steps are less affected by imprecise explanations. In addition, our demonstrations in the prompt better resemble the more complex steps. This points to the fact that further refinement of the selection of demonstrations could further improve performance. In this case, there appears to be a very limited effect when altering the history available to the LLM.


\begin{table}[]
\small
\caption{Avg. ratings for different configurations. The dimensions are as specified in Table \ref{tab:translation-eval}}
\centering
\resizebox{\columnwidth}{!}{%
\begin{tabular}{@{}llllllll@{}}
\toprule
                                  & \textbf{History} & \textbf{A}    & \textbf{B}    & \textbf{C}    & \textbf{D}    & \textbf{E}    & \textbf{F}    \\ \midrule
\multirow{2}{*}{\textbf{Trivial Steps}} & Full History          & 2.96 & 4.02 & 4.24 & 2.52 & 4.24 & 3.82 \\ \cmidrule(l){2-8} 
                                  & Filtered History      & 3.34 & 3.88 & 3.94 & 2.4  & 4.3  & 4.22 \\ \midrule
\multirow{2}{*}{\textbf{Complex Steps}} & Full History          & 4.48 & 4.74 & 4.44 & 1.42 & 4.34 & 3.12 \\ \cmidrule(l){2-8} 
                                  & Filtered History      & 4.4  & 4.4  & 4.52 & 1.64 & 4.34 & 3.28 \\ \bottomrule
\end{tabular}}
\label{tab:average-rating}
\end{table}

\subsection{Limitations}
\label{sec:limitations}

First, while our initial results are promising, we have not yet developed a complete system; full experimental results, including human studies, are forthcoming. 
Second, our experiments are limited to Hitori puzzles.  While we have achieved promising preliminary results with other puzzles like Sudoku, Futoshiki, and Star Battle (see Appendix), it remains an open question how well these results generalize to other domains.  We expect extensions and adaptations of the proof staging approach will be called for, when more than two styles of explanations become necessary.
Third, we have evaluated our approach using only the Z3 solver; exploring its effectiveness across other SAT and SMT solvers is left for future work. 
Finally, although Hitori is NP-complete, the exact complexity of the proof staging problem itself remains unknown.

\section{Related Work and Conclusion}
\label{sec:conclusion}
LLMs have been shown to struggle to solve puzzles that require complex reasoning~\cite{giadikiaroglou2024puzzlesolvingusingreasoning}. In response to these challenges, methods that combine symbolic solvers and LLMs have recently received increasing attention~\cite{mittal2024puzzlebench, Jiang2022DraftSA, Xin2024DeepSeekProverAT}. Common approaches include using LLMs to translate problem specifications and proof sketches into formal programs, then offload them to standard solvers~\cite{jiang2023draft,mittal2024puzzlebench}, or generating large-scale synthetic formal proof data to fine-tune LLMs~\cite{Xin2024DeepSeekProverAT}. However, significant less attention has been paid to the problem of whether LLMs can produce useful explanations (in the form of reasoning steps) for proof objects. 

A significant body of research has addressed explanations in the formal verification and planning domains. Early work ~\cite{jin2002fate,groce2003went} focused on generating human-understandable explanations for model violations through counterexamples. This was later refined through causality-based reasoning~\cite{beer2012explaining} and abstraction techniques to guide the analysis of error traces~\cite{nanshi2013using}. While these methods effectively explain isolated violations or short decision sequences, they are less suited for complex, multi-step processes involving interconnected constraints.
More recently, research attention has shifted to explainability in policy synthesis for stochastic systems~\cite{brazdil2015counterexample,azeem2024explainable}. 
However, these approaches primarily rely on symbolic reasoning and lack mechanisms for producing natural language explanations.

Our work advances these threads by proposing a neurosymbolic approach that combines decision procedures with LLMs to explain complex sequences of decisions. Using Hitori puzzles as a case study, we demonstrate how local constraints can be formally explained through short resolution proofs, while global connectivity constraints benefit from visual and language-based explanations. This hybrid approach leverages the strengths of modern SAT solvers~\cite{Moura11, Barbos22, Biere24} along with the NLP capabilities of LLMs, providing a flexible and effective framework for human-centered explanations.





\bibliography{Sections/local,Sections/References}

\begin{thebibliography}{36}
\providecommand{\natexlab}[1]{#1}

\bibitem[{Anderson and Bledsoe(1970)}]{Anders70}
Anderson, R.; and Bledsoe, W.~W. 1970.
\newblock A Linear Format for Resolution With Merging and a New Technique for Establishing Completeness.
\newblock \emph{Journal of the Association for Computing Machinery}, 17(3): 525--534.

\bibitem[{Azeem et~al.(2024)Azeem, Chakraborty, Kanav, and Kretinsky}]{azeem2024explainable}
Azeem, M.; Chakraborty, D.; Kanav, S.; and Kretinsky, J. 2024.
\newblock Explainable Finite-Memory Policies for Partially Observable Markov Decision Processes.
\newblock \emph{arXiv preprint arXiv:2411.13365}.

\bibitem[{Barbosa et~al.(2022)Barbosa, Barrett, Brain, Kremer, Lachnitt, Mann, Mohamed, Mohamed, Niemetz, N{\"{o}}tzli, Ozdemir, Preiner, Reynolds, Sheng, Tinelli, and Zohar}]{Barbos22}
Barbosa, H.; Barrett, C.~W.; Brain, M.; Kremer, G.; Lachnitt, H.; Mann, M.; Mohamed, A.; Mohamed, M.; Niemetz, A.; N{\"{o}}tzli, A.; Ozdemir, A.; Preiner, M.; Reynolds, A.; Sheng, Y.; Tinelli, C.; and Zohar, Y. 2022.
\newblock cvc5: {A} Versatile and Industrial-Strength {SMT} Solver.
\newblock In \emph{Tools and Algorithms for the Construction and Analysis of Systems {TACAS} 2022}, 415--442.
\newblock LNCS 13243.

\bibitem[{Beer et~al.(2012)Beer, Ben-David, Chockler, Orni, and Trefler}]{beer2012explaining}
Beer, I.; Ben-David, S.; Chockler, H.; Orni, A.; and Trefler, R. 2012.
\newblock Explaining counterexamples using causality.
\newblock \emph{Formal Methods in System Design}, 40: 20--40.

\bibitem[{Biere et~al.(2024)Biere, Faller, Fazekas, Fleury, Froleyks, and Pollitt}]{Biere24}
Biere, A.; Faller, T.; Fazekas, K.; Fleury, M.; Froleyks, N.; and Pollitt, F. 2024.
\newblock {CaDiCaL 2.0}.
\newblock In \emph{Computer Aided Verification (CAV)}, 133--152.
\newblock LNCS 14681.

\bibitem[{Br{\'a}zdil et~al.(2015)Br{\'a}zdil, Chatterjee, Chmel{\'\i}k, Fellner, and K{\v{r}}et{\'\i}nsk{\`y}}]{brazdil2015counterexample}
Br{\'a}zdil, T.; Chatterjee, K.; Chmel{\'\i}k, M.; Fellner, A.; and K{\v{r}}et{\'\i}nsk{\`y}, J. 2015.
\newblock Counterexample explanation by learning small strategies in Markov decision processes.
\newblock In \emph{Computer Aided Verification: 27th International Conference, CAV 2015, San Francisco, CA, USA, July 18-24, 2015, Proceedings, Part I}, 158--177. Springer.

\bibitem[{Brock-Nannestad(2018)}]{Brock18}
Brock-Nannestad, T. 2018.
\newblock Space-efficient acyclicity constraints: A declarative pearl.
\newblock \emph{Science of Computer Programming}, 164: 66--81.

\bibitem[{Brown et~al.(2020)Brown, Mann, Ryder, Subbiah, Kaplan, Dhariwal, Neelakantan, Shyam, Sastry, Askell, Agarwal, Herbert-Voss, Krueger, Henighan, Child, Ramesh, Ziegler, Wu, Winter, Hesse, Chen, Sigler, Litwin, Gray, Chess, Clark, Berner, McCandlish, Radford, Sutskever, and Amodei}]{NEURIPS2020_1457c0d6}
Brown, T.; Mann, B.; Ryder, N.; Subbiah, M.; Kaplan, J.~D.; Dhariwal, P.; Neelakantan, A.; Shyam, P.; Sastry, G.; Askell, A.; Agarwal, S.; Herbert-Voss, A.; Krueger, G.; Henighan, T.; Child, R.; Ramesh, A.; Ziegler, D.; Wu, J.; Winter, C.; Hesse, C.; Chen, M.; Sigler, E.; Litwin, M.; Gray, S.; Chess, B.; Clark, J.; Berner, C.; McCandlish, S.; Radford, A.; Sutskever, I.; and Amodei, D. 2020.
\newblock Language Models are Few-Shot Learners.
\newblock In Larochelle, H.; Ranzato, M.; Hadsell, R.; Balcan, M.; and Lin, H., eds., \emph{Advances in Neural Information Processing Systems}, volume~33, 1877--1901. Curran Associates, Inc.

\bibitem[{Chen et~al.(2024)Chen, Phang, Parrish, Padmakumar, Zhao, Bowman, and Cho}]{chen2024two}
Chen, A.; Phang, J.; Parrish, A.; Padmakumar, V.; Zhao, C.; Bowman, S.~R.; and Cho, K. 2024.
\newblock Two Failures of Self-Consistency in the Multi-Step Reasoning of {LLM}s.
\newblock \emph{Transactions on Machine Learning Research}.

\bibitem[{Chkirbene et~al.(2024)Chkirbene, Hamila, Gouissem, and Devrim}]{Chkirbene2024LargeLM}
Chkirbene, Z.; Hamila, R.; Gouissem, A.; and Devrim, U. 2024.
\newblock Large Language Models (LLM) in Industry: A Survey of Applications, Challenges, and Trends.
\newblock \emph{2024 IEEE 21st International Conference on Smart Communities: Improving Quality of Life using AI, Robotics and IoT (HONET)}, 229--234.

\bibitem[{Davis and Putnam(1960)}]{Davis60}
Davis, M.; and Putnam, H. 1960.
\newblock A Computing Procedure for Quantification Theory.
\newblock \emph{Journal of the ACM}, 7(3): 201--215.

\bibitem[{de~Moura and Bj{\o}rner(2008)}]{Moura08b}
de~Moura, L.; and Bj{\o}rner, N. 2008.
\newblock Proofs and Refutations, and {Z3}.
\newblock In \emph{LPAR Workshops}, 123--132.

\bibitem[{de~Moura and Bj{\o}rner(2011)}]{Moura11}
de~Moura, L.; and Bj{\o}rner, N. 2011.
\newblock Satisfiability Modulo Theories: Introduction and Applications.
\newblock \emph{Communications of the ACM}, 54(9): 69--77.

\bibitem[{DeepSeek-AI et~al.(2025)DeepSeek-AI, Guo, Yang, Zhang, Song, Zhang, Xu, Zhu, Ma, Wang, Bi, Zhang, Yu, Wu, Wu, Gou, Shao, Li, Gao, Liu, Xue, Wang, Wu, Feng, Lu, Zhao, Deng, Zhang, Ruan, Dai, Chen, Ji, Li, Lin, Dai, Luo, Hao, Chen, Li, Zhang, Bao, Xu, Wang, Ding, Xin, Gao, Qu, Li, Guo, Li, Wang, Chen, Yuan, Qiu, Li, Cai, Ni, Liang, Chen, Dong, Hu, Gao, Guan, Huang, Yu, Wang, Zhang, Zhao, Wang, Zhang, Xu, Xia, Zhang, Zhang, Tang, Li, Wang, Li, Tian, Huang, Zhang, Wang, Chen, Du, Ge, Zhang, Pan, Wang, Chen, Jin, Chen, Lu, Zhou, Chen, Ye, Wang, Yu, Zhou, Pan, Li, Zhou, Wu, Ye, Yun, Pei, Sun, Wang, Zeng, Zhao, Liu, Liang, Gao, Yu, Zhang, Xiao, An, Liu, Wang, Chen, Nie, Cheng, Liu, Xie, Liu, Yang, Li, Su, Lin, Li, Jin, Shen, Chen, Sun, Wang, Song, Zhou, Wang, Shan, Li, Wang, Wei, Zhang, Xu, Li, Zhao, Sun, Wang, Yu, Zhang, Shi, Xiong, He, Piao, Wang, Tan, Ma, Liu, Guo, Ou, Wang, Gong, Zou, He, Xiong, Luo, You, Liu, Zhou, Zhu, Xu, Huang, Li, Zheng, Zhu, Ma, Tang, Zha, Yan, Ren, Ren, Sha, Fu, Xu, Xie, Zhang,
  Hao, Ma, Yan, Wu, Gu, Zhu, Liu, Li, Xie, Song, Pan, Huang, Xu, Zhang, and Zhang}]{deepseekai2025deepseekr1incentivizingreasoningcapability}
DeepSeek-AI; Guo, D.; Yang, D.; Zhang, H.; Song, J.; Zhang, R.; Xu, R.; Zhu, Q.; Ma, S.; Wang, P.; Bi, X.; Zhang, X.; Yu, X.; Wu, Y.; Wu, Z.~F.; Gou, Z.; Shao, Z.; Li, Z.; Gao, Z.; Liu, A.; Xue, B.; Wang, B.; Wu, B.; Feng, B.; Lu, C.; Zhao, C.; Deng, C.; Zhang, C.; Ruan, C.; Dai, D.; Chen, D.; Ji, D.; Li, E.; Lin, F.; Dai, F.; Luo, F.; Hao, G.; Chen, G.; Li, G.; Zhang, H.; Bao, H.; Xu, H.; Wang, H.; Ding, H.; Xin, H.; Gao, H.; Qu, H.; Li, H.; Guo, J.; Li, J.; Wang, J.; Chen, J.; Yuan, J.; Qiu, J.; Li, J.; Cai, J.~L.; Ni, J.; Liang, J.; Chen, J.; Dong, K.; Hu, K.; Gao, K.; Guan, K.; Huang, K.; Yu, K.; Wang, L.; Zhang, L.; Zhao, L.; Wang, L.; Zhang, L.; Xu, L.; Xia, L.; Zhang, M.; Zhang, M.; Tang, M.; Li, M.; Wang, M.; Li, M.; Tian, N.; Huang, P.; Zhang, P.; Wang, Q.; Chen, Q.; Du, Q.; Ge, R.; Zhang, R.; Pan, R.; Wang, R.; Chen, R.~J.; Jin, R.~L.; Chen, R.; Lu, S.; Zhou, S.; Chen, S.; Ye, S.; Wang, S.; Yu, S.; Zhou, S.; Pan, S.; Li, S.~S.; Zhou, S.; Wu, S.; Ye, S.; Yun, T.; Pei, T.; Sun, T.; Wang, T.; Zeng, W.;
  Zhao, W.; Liu, W.; Liang, W.; Gao, W.; Yu, W.; Zhang, W.; Xiao, W.~L.; An, W.; Liu, X.; Wang, X.; Chen, X.; Nie, X.; Cheng, X.; Liu, X.; Xie, X.; Liu, X.; Yang, X.; Li, X.; Su, X.; Lin, X.; Li, X.~Q.; Jin, X.; Shen, X.; Chen, X.; Sun, X.; Wang, X.; Song, X.; Zhou, X.; Wang, X.; Shan, X.; Li, Y.~K.; Wang, Y.~Q.; Wei, Y.~X.; Zhang, Y.; Xu, Y.; Li, Y.; Zhao, Y.; Sun, Y.; Wang, Y.; Yu, Y.; Zhang, Y.; Shi, Y.; Xiong, Y.; He, Y.; Piao, Y.; Wang, Y.; Tan, Y.; Ma, Y.; Liu, Y.; Guo, Y.; Ou, Y.; Wang, Y.; Gong, Y.; Zou, Y.; He, Y.; Xiong, Y.; Luo, Y.; You, Y.; Liu, Y.; Zhou, Y.; Zhu, Y.~X.; Xu, Y.; Huang, Y.; Li, Y.; Zheng, Y.; Zhu, Y.; Ma, Y.; Tang, Y.; Zha, Y.; Yan, Y.; Ren, Z.~Z.; Ren, Z.; Sha, Z.; Fu, Z.; Xu, Z.; Xie, Z.; Zhang, Z.; Hao, Z.; Ma, Z.; Yan, Z.; Wu, Z.; Gu, Z.; Zhu, Z.; Liu, Z.; Li, Z.; Xie, Z.; Song, Z.; Pan, Z.; Huang, Z.; Xu, Z.; Zhang, Z.; and Zhang, Z. 2025.
\newblock DeepSeek-R1: Incentivizing Reasoning Capability in LLMs via Reinforcement Learning.
\newblock arXiv:2501.12948.

\bibitem[{Giadikiaroglou et~al.(2024)Giadikiaroglou, Lymperaiou, Filandrianos, and Stamou}]{giadikiaroglou2024puzzlesolvingusingreasoning}
Giadikiaroglou, P.; Lymperaiou, M.; Filandrianos, G.; and Stamou, G. 2024.
\newblock Puzzle Solving using Reasoning of Large Language Models: A Survey.
\newblock arXiv:2402.11291.

\bibitem[{Groce and Visser(2003)}]{groce2003went}
Groce, A.; and Visser, W. 2003.
\newblock What went wrong: Explaining counterexamples.
\newblock In \emph{International SPIN Workshop on Model Checking of Software}, 121--136. Springer.

\bibitem[{Hearn and Demaine(2009)}]{Hearn09}
Hearn, R.~A.; and Demaine, E.~D. 2009.
\newblock \emph{Games, Puzzles, and Computation}.
\newblock A K Peters.

\bibitem[{Heule and Biere(2015)}]{Heule15}
Heule, M. J.~H.; and Biere, A. 2015.
\newblock Proofs for Satisfiability Problems.
\newblock In \emph{All About Proofs, Proofs for All (APPA)}.

\bibitem[{Heule, Kullmann, and Marek(2016)}]{Heule16}
Heule, M. J.~H.; Kullmann, O.; and Marek, V.~W. 2016.
\newblock Solving and Verifying the boolean {Pythagorean} Triples problem via Cube-and-Conquer.
\newblock In \emph{Theory and Applications of Satisfiability Testing (SAT 2016)}, 228--245.
\newblock LNCS 9710.

\bibitem[{Hopcroft and Tarjan(1973)}]{Hopcro73}
Hopcroft, J.; and Tarjan, R. 1973.
\newblock Algorithm 447: efficient algorithms for graph manipulation.
\newblock \emph{Communications of the ACM}, 16(6): 372--378.

\bibitem[{Janota, Lynce, and Marques-Silva(2015)}]{Janota15}
Janota, M.; Lynce, I.; and Marques-Silva, J. 2015.
\newblock Algorithms for computing backbones of propositonal formulae.
\newblock \emph{AI Communications}, 28: 161--177.

\bibitem[{Jiang et~al.(2023)Jiang, Welleck, Zhou, Lacroix, Liu, Li, Jamnik, Lample, and Wu}]{jiang2023draft}
Jiang, A.~Q.; Welleck, S.; Zhou, J.~P.; Lacroix, T.; Liu, J.; Li, W.; Jamnik, M.; Lample, G.; and Wu, Y. 2023.
\newblock Draft, Sketch, and Prove: Guiding Formal Theorem Provers with Informal Proofs.
\newblock In \emph{The Eleventh International Conference on Learning Representations}.

\bibitem[{Jiang et~al.(2022)Jiang, Welleck, Zhou, Li, Liu, Jamnik, Lacroix, Wu, and Lample}]{Jiang2022DraftSA}
Jiang, A.~Q.; Welleck, S.; Zhou, J.~P.; Li, W.; Liu, J.; Jamnik, M.; Lacroix, T.; Wu, Y.; and Lample, G. 2022.
\newblock Draft, Sketch, and Prove: Guiding Formal Theorem Provers with Informal Proofs.
\newblock \emph{ArXiv}, abs/2210.12283.

\bibitem[{Jin, Ravi, and Somenzi(2002)}]{jin2002fate}
Jin, H.; Ravi, K.; and Somenzi, F. 2002.
\newblock Fate and FreeWill in error traces.
\newblock In \emph{International Conference on Tools and Algorithms for the Construction and Analysis of Systems}, 445--459. Springer.

\bibitem[{K{\"a}mmer et~al.(2024)K{\"a}mmer, Hautz, Krummrey, Sauter, Penders, Birrenbach, and Bienefeld}]{Kammere087469}
K{\"a}mmer, J.~E.; Hautz, W.~E.; Krummrey, G.; Sauter, T.~C.; Penders, D.; Birrenbach, T.; and Bienefeld, N. 2024.
\newblock Effects of interacting with a large language model compared with a human coach on the clinical diagnostic process and outcomes among fourth-year medical students: study protocol for a prospective, randomised experiment using patient vignettes.
\newblock \emph{BMJ Open}, 14(7).

\bibitem[{Kiesl, Rebola-Pardo, and Heule(2018)}]{Kiesl18}
Kiesl, B.; Rebola-Pardo, A.; and Heule, M. J.~H. 2018.
\newblock Extended Resolution Simulates {DRAT}.
\newblock In \emph{International Joint Conference on Automated Reasoning}, 516--531.

\bibitem[{Kilby et~al.(2005)Kilby, Slaney, Thi{\'e}baux, and Walsh}]{Kilby05}
Kilby, P.; Slaney, J.; Thi{\'e}baux, S.; and Walsh, T. 2005.
\newblock Backbones and Backdoors in Satisfiability.
\newblock In \emph{AAAI-05}, 1368--1373.

\bibitem[{Kim et~al.(2025)Kim, Choe, Yoo, Chowdhury, and Seo}]{kim2025toolsunderstandingheavyusers}
Kim, E.; Choe, K.; Yoo, M.; Chowdhury, S.~S.; and Seo, J. 2025.
\newblock Beyond Tools: Understanding How Heavy Users Integrate LLMs into Everyday Tasks and Decision-Making.
\newblock arXiv:2502.15395.

\bibitem[{Knuth(2023)}]{Knuth-v4b}
Knuth, D.~E. 2023.
\newblock \emph{The Art of Computer Programming}, volume~4B.
\newblock Addison Wesley.

\bibitem[{Lewis et~al.(2020)Lewis, Perez, Piktus, Petroni, Karpukhin, Goyal, K\"{u}ttler, Lewis, Yih, Rockt\"{a}schel, Riedel, and Kiela}]{NEURIPS2020_6b493230}
Lewis, P.; Perez, E.; Piktus, A.; Petroni, F.; Karpukhin, V.; Goyal, N.; K\"{u}ttler, H.; Lewis, M.; Yih, W.-t.; Rockt\"{a}schel, T.; Riedel, S.; and Kiela, D. 2020.
\newblock Retrieval-Augmented Generation for Knowledge-Intensive NLP Tasks.
\newblock In Larochelle, H.; Ranzato, M.; Hadsell, R.; Balcan, M.; and Lin, H., eds., \emph{Advances in Neural Information Processing Systems}, volume~33, 9459--9474. Curran Associates, Inc.

\bibitem[{{Marques-Silva} and Sakallah(1999)}]{Silva99c}
{Marques-Silva}, J.~P.; and Sakallah, K.~A. 1999.
\newblock {GRASP}: A Search Algorithm for Propositional Satisfiability.
\newblock \emph{IEEE Transactions on Computers}, 48(5): 506--521.

\bibitem[{Mittal et~al.(2024)Mittal, Kartik, Singla et~al.}]{mittal2024puzzlebench}
Mittal, C.; Kartik, K.; Singla, P.; et~al. 2024.
\newblock PuzzleBench: Can {LLMs} Solve Challenging First-Order Combinatorial Reasoning Problems?
\newblock \emph{arXiv preprint arXiv:2402.02611}.

\bibitem[{Nanshi and Somenzi(2013)}]{nanshi2013using}
Nanshi, K.; and Somenzi, F. 2013.
\newblock Using abstraction to guide the search for long error traces.
\newblock \emph{IEEE Transactions on Computer-Aided Design of Integrated Circuits and Systems}, 32(3): 453--466.

\bibitem[{Robinson(1965)}]{Robins65}
Robinson, J.~A. 1965.
\newblock A Machine-Oriented Logic Based on the Resolution Principle.
\newblock \emph{Journal of the Association for Computing Machinery}, 12(1): 23--41.

\bibitem[{Turpin et~al.(2023)Turpin, Michael, Perez, and Bowman}]{turpin2023language}
Turpin, M.; Michael, J.; Perez, E.; and Bowman, S.~R. 2023.
\newblock Language Models Don't Always Say What They Think: Unfaithful Explanations in Chain-of-Thought Prompting.
\newblock In \emph{Thirty-seventh Conference on Neural Information Processing Systems}.

\bibitem[{Xin et~al.(2024)Xin, Guo, Shao, Ren, Zhu, Liu), Ruan, Li, and Liang}]{Xin2024DeepSeekProverAT}
Xin, H.; Guo, D.; Shao, Z.; Ren, Z.; Zhu, Q.; Liu), B. L.~B.; Ruan, C.; Li, W.; and Liang, X. 2024.
\newblock DeepSeek-Prover: Advancing Theorem Proving in {LLMs} through Large-Scale Synthetic Data.
\newblock \emph{ArXiv}, abs/2405.14333.

\end{thebibliography}




\newpage
\appendix
\onecolumn
\setcounter{secnumdepth}{2} 

\section{Details of the Encoding}
\label{app:hitori-encoding}

\subsection{Acyclicity Constraints}
\label{sec:preventing-loops}

The challenge in encoding Hitori for a SAT/SMT solver lies in the
connectivity constraints.  The general strategy is to add constraints
whose satisfaction implies the existence of a spanning tree of all
unshaded squares.  Each square---except a designated root---is
equipped with a \emph{parent link}, which the solver causes to point to
the square's parent in the spanning tree.
These links should form no cycles.  We could let each link be a non-negative
integer-valued variable.  The root would be numbered $0$, while every
other unshaded square would have to point to an unshaded square with a
smaller integer value.  This, however, is not very efficient.
Instead, we adopt an approach based on \cite{Brock18}, whose main ideas we summarize here.  (More details and proofs are found in \cite{Brock18}.)

Consider the cycles below. As we go around them in \emph{clockwise} fashion,  we encounter four and eight turns, respectively.
\begin{center}
  \begin{tikzpicture}[scale=0.4]
    \begin{scope}
      \draw[ultra thin, gray] (0,0) grid (5,5);
      \fill[orange!50] (1.5,3.5) circle[radius=0.25cm];
      \draw[very thick,blue,->] (1.5,1.5) -- ++(0,2) -- ++(2,0) --
      ++(0,-2) -- ++(-2,0);
    \end{scope}
    \begin{scope}[xshift=6.5cm]
      \fill[orange!50] (1.5,3.5) circle[radius=0.25cm];
      \fill[orange!50] (3.5,4.5) circle[radius=0.25cm];
      \fill[green!50] (3.25,3.25) rectangle (3.75,3.75);
      \draw[very thick,blue,->] (1.5,1.5) -- ++(0,2) -- ++(2,0) --
      ++(0,1) -- ++(1,0) -- ++(0,-2) -- ++(-1,0) -- ++(0,-1) --
      ++(-2,0);
      \draw[ultra thin, gray] (0,0) grid (5,5); 
    \end{scope}
  \end{tikzpicture}
\end{center}
We identify these turns by the directions
before and after the turn.  We focus on two types of turns: up-then-right (marked by orange circles) and right-then-up (marked by green squares).  We call these turns \emph{notable}.  In the cycle on the left, there is one up-then-right turn and no right-then-up
turns.  In the cycle on the right, there are two turns of the first
type and one turn of the second.  In general, in every cycle there is always one more up-then-right turn than there are right-then-up turns.  So, the number
of notable turns is always odd.  
Moreover, the difference between the numbers of notable turns of the two types is always $0$ or $1$ along the cycle and is $1$ over one lap.  Hence, we can track this difference with just one bit.

To convert this idea into constraints for a SAT solver, we associate one Boolean variable to each square of the grid.  We impose the constraint that, if the cycle turns up-then-right or
right-then-up in that square, the turn-tracking variable of this square
must have opposite value to that of its predecessor along the path.
In the remaining $10$ ways in which the cycle goes through a square, the
turn-tracking variable must have the same value as the variable from
the predecessor square.  The turn-tracking variables of the shaded squares
may take arbitrary values.

The chain of constraints around a cycle forces the turn tracking bit to change an odd number of times.
This means that the parity constraints around a cycle are unsatisfiable.  Hence, there are no cycles.

\subsection{The Encoding}

We detail how to model a Hitori puzzle for a SAT solver.  For an $m \times n$ grid, with square $(0,0)$ in the top-left corner, we associate six Boolean variables to each square of the grid.
\begin{itemize}[leftmargin=*]
\item $c_{i,j}$: if it is true, then square $(i,j)$ is unshaded;
  otherwise, it is shaded.
\item $h_{i,j,1}$ and $h_{i,j,0}$: they encode the horizontal parent
  link according to the following scheme:
  \begin{itemize}
  \item $(\bot,\bot)$: no horizontal link
  \item $(\bot,\top)$: the horizontal link points right
  \item $(\top,\top)$: the horizontal link points left
  \item $(\top,\top)$: forbidden
  \end{itemize}
\item $v_{i,j,1}$ and $v_{i,j,0}$: they encode the vertical parent
  link according to the following scheme:
  \begin{itemize}
  \item $(\bot,\bot)$: no vertical link
  \item $(\bot,\top)$: the vertical link points up
  \item $(\top,\bot)$: the vertical link points down
  \item $(\top,\top)$: forbidden
  \end{itemize}
\item $p_{i,j}$: the turn parity of square $(i,j)$ that is discussed
  in Section~\ref{sec:preventing-loops}.
\end{itemize}
We also need a single Boolean variable, $r$, to
select the root of the tree.  Let's start with the constraints on the
$c$ variables.
\begin{itemize}[leftmargin=*]
\item If squares $(i,j)$ and $(k,\ell)$ in the same row (i.e., $i=k$)
  or column (i.e., $j = \ell$) of the grid contain the same symbol, at
  least one of them must be shaded:
  $\neg c_{i,j} \vee \neg c_{k,\ell}$.
\item No two shaded squares should share a side: if squares $(i,j)$
  and $(k,\ell)$ are neighbors, $c_{i,j} \vee c_{k,\ell}$.
\end{itemize}
If these constraints on the $c$ variables are satisfied, but the
constraints on the remaining variables are ignored, we get a
\emph{weak} solution.  In a weak solution, the unshaded squares may
form more than one connected region.

For the constraints on the $h$ and $v$ variables, we need
to distinguish three cases: The shaded squares, the root of the spanning tree, and the other unshaded squares.  
The $h$, $v$, and $p$ variables of the shaded squares are unconstrained.
Every unshaded square that is not the root must have exactly one outgoing edge, pointing to its parent in
the tree.  In addition, if the horizontal parent pointer of unshaded
square $(i,j)$ points left, then $j > 0$ should hold and $c_{i,j-1}$
should be $\top$.  Likewise for the other three directions.  Finally, two unshaded squares should not point at each
other.  (These ``short loops'' are not prevented by the $p$ variables.)

The root of the tree can be confined to the first two squares of the first row,  because no two adjacent squares can both be shaded.  One Boolean variable, $r$, is used to let the SAT solver decide where the root will be.  We stipulate that $r$ is true if the root is $(0,1)$ and false if the root is $(0,0)$.  Then, if $r$ is true, then $c_{0,1}$ is true, and if $r$ is false, then $c_{0,0}$ is true.

As discussed in Appendix~\ref{sec:preventing-loops}, the variable $p_{i,j}$, should switch value whenever the loop takes an up-then-right or right-then-up turn.  The particular values of the $p$ variables are not important: only their relations to the neighboring variables if they belong to unshaded cells.

The constraint on the $p$ variables at square ($i,j)$ have the
following form: ``If there is an incoming edge from above, and an
outgoing edge to the left, then $p_{i,j}$ and $p_{i-1,j}$ should be
the same.  There are $12$ possible
combinations of incoming and outgoing edges to consider, some of which may not be possible near the boundaries of the grid.  As an example, consider this fragment of grid:
\begin{center}
  \begin{tikzpicture}[scale=0.8,transform shape,
    loop/.style={rounded corners,blue,ultra thick},
    emptyc/.style={draw=black,semithick,fill=yellow!5}]
    \fill[yellow!5] (0,0) rectangle +(3,3);
    \fill[blue] (1.55,0.05) rectangle +(1.4,1.4);
    \draw[thin,step=1.5cm] (0,0) grid (3,3);
    \draw[loop,->] (0.75, 0.85) |- (3.4,2.25);
    \node (ind-i-j) at (0.75,2.7) {\small$(i,j)$};
    \node (ind-i-jp1) at (2.25,2.7) {\small$(i,j+1)$};
    \node (ind-ip1-j) at (0.75,0.5) {\small$(i+1,j)$};
  \end{tikzpicture}
\end{center}
For these values of the $c$, $h$, and $v$ variables, the constraint on
the $p$ variables should be,
\begin{equation*}
  (p_{i,j} \neq p_{i+1,j}) \wedge (p_{i,j+1} = p_{i,j}) \enspace.
\end{equation*}
Suppose $\operatorname{up}(i,j)$ is a function of the $h$ and $v$
variables that is true if, and only if, the arrow from square $(i,j)$
points up.  Likewise, for 
$\operatorname{right}(i,j)$.
We need the constraints,
\begin{gather*}
  (\operatorname{up}(i+1,j) \wedge \operatorname{right}(i,j))
  \rightarrow  p_{i,j} \neq p_{i+1,j} \\
  (\operatorname{right}(i,j) \wedge \operatorname{right}(i,j+1))
  \rightarrow p_{i,j+1} = p_{i,j} \enspace.
\end{gather*}

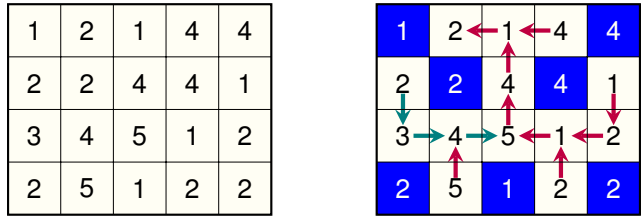
\begin{figure}
  \centering
  \begin{tikzpicture}[scale=0.7,transform shape,
    treeedge/.style={-stealth,ultra thick}]
    \sffamily\sansmath\LARGE
    \colorlet{shaded}{blue}
    \colorlet{background}{yellow!5!white}
    \begin{scope}
      \fill[background] (0,0) rectangle (5,4);
      \node[black] at (0.5,3.5) {1};
      \node[black] at (1.5,3.5) {2};
      \node[black] at (2.5,3.5) {1};
      \node[black] at (3.5,3.5) {4};
      \node[black] at (4.5,3.5) {4};
      \node[black] at (0.5,2.5) {2};
      \node[black] at (1.5,2.5) {2};
      \node[black] at (2.5,2.5) {4};
      \node[black] at (3.5,2.5) {4};
      \node[black] at (4.5,2.5) {1};
      \node[black] at (0.5,1.5) {3};
      \node[black] at (1.5,1.5) {4};
      \node[black] at (2.5,1.5) {5};
      \node[black] at (3.5,1.5) {1};
      \node[black] at (4.5,1.5) {2};
      \node[black] at (0.5,0.5) {2};
      \node[black] at (1.5,0.5) {5};
      \node[black] at (2.5,0.5) {1};
      \node[black] at (3.5,0.5) {2};
      \node[black] at (4.5,0.5) {2};
      \draw[step=1cm,ultra thin] (0,0) grid (5,4);
      \draw[thick] (0,0) rectangle (5,4);
    \end{scope}
    \begin{scope}[xshift=7cm]
      \fill[background] (0,0) rectangle (5,4);
      \fill[shaded] (0,3) rectangle +(1,1);
      \node[background] at (0.5,3.5) {1};
      \node[black] at (1.5,3.5) {2};
      \node[black] at (2.5,3.5) {1};
      \node[black] at (3.5,3.5) {4};
      \fill[shaded] (4,3) rectangle +(1,1);
      \node[background] at (4.5,3.5) {4};
      \node[black] at (0.5,2.5) {2};
      \fill[shaded] (1,2) rectangle +(1,1);
      \node[background] at (1.5,2.5) {2};
      \node[black] at (2.5,2.5) {4};
      \fill[shaded] (3,2) rectangle +(1,1);
      \node[background] at (3.5,2.5) {4};
      \node[black] at (4.5,2.5) {1};
      \node[black] at (0.5,1.5) {3};
      \node[black] at (1.5,1.5) {4};
      \node[black] at (2.5,1.5) {5};
      \node[black] at (3.5,1.5) {1};
      \node[black] at (4.5,1.5) {2};
      \fill[shaded] (0,0) rectangle +(1,1);
      \node[background] at (0.5,0.5) {2};
      \node[black] at (1.5,0.5) {5};
      \fill[shaded] (2,0) rectangle +(1,1);
      \node[background] at (2.5,0.5) {1};
      \node[black] at (3.5,0.5) {2};
      \fill[shaded] (4,0) rectangle +(1,1);
      \node[background] at (4.5,0.5) {2};
      \draw[treeedge,purple] (2.3,3.5) -- +(-0.6,0.0);
      \draw[treeedge,purple] (3.3,3.5) -- +(-0.6,0.0);
      \draw[treeedge,teal] (0.5,2.3) -- +(0.0,-0.6);
      \draw[treeedge,purple] (2.5,2.7) -- +(0.0,0.6);
      \draw[treeedge,purple] (4.5,2.3) -- +(0.0,-0.6);
      \draw[treeedge,teal] (0.7,1.5) -- +(0.6,0.0);
      \draw[treeedge,teal] (1.7,1.5) -- +(0.6,0.0);
      \draw[treeedge,purple] (2.5,1.7) -- +(0.0,0.6);
      \draw[treeedge,purple] (3.3,1.5) -- +(-0.6,0.0);
      \draw[treeedge,purple] (4.3,1.5) -- +(-0.6,0.0);
      \draw[treeedge,purple] (1.5,0.7) -- +(0.0,0.6);
      \draw[treeedge,purple] (3.5,0.7) -- +(0.0,0.6);
      \draw[step=1cm,ultra thin] (0,0) grid (5,4);
      \draw[thick] (0,0) rectangle (5,4);
    \end{scope}
  \end{tikzpicture}
  \caption{Possible values of the $p$ variables of a puzzle.  The root
    of the spanning tree is Square~$(0,1)$.}
  \label{fig:hitori-p-values}
\end{figure}
For a puzzle with $m=4$, $n=5$, Figure~\ref{fig:hitori-p-values} shows
a possible spanning tree.  The arrows describe the values of the $h$
and $v$ variables.  Each arrow is colored according the value of the
square's $p$ variable: red means true and teal means false.

\section{Algorithm}
\label{sec:algorithm}

The pseudocode for our proof staging algorithm is shown in Algorithm~\ref{algo:main}.
\begin{algorithm}[t]
\caption{\textsc{ExplainHitoriSolution}$(f, b)$}
\label{algo:main}
        subgoals $\leftarrow$ \textsc{ExtractSubgoalsFromBlockingClause}$(b)$\;
        \While{subgoals are not empty}{
            \ForEach{subgoal in subgoals}{
                weak\_proof $\leftarrow$ \textsc{WeakSolver}$(f \wedge \neg \text{subgoal})$\;
                \eIf{weak\_proof exists}{
                    Rank subgoal by \textsc{Size}$(\text{weak\_proof})$\;
                }{
                    Mark subgoal as requiring strong proof\;
                }
            }
            next\_subgoal $\leftarrow$ \textsc{SelectSubgoalWithSmallestProof}$()$\;
            \eIf{next\_subgoal requires strong proof}{
                \textsc{DisplayProofByPicture}$(\text{next\_subgoal}, f)$\;
            }{
                \textsc{ExplainProofInText}$(\text{next\_subgoal}, \text{weak\_proof})$\;
            }
            \textsc{ApplyDeduction}$(f, \text{next\_subgoal})$\;
            Remove next\_subgoal from subgoals\;
        }
\end{algorithm}
The inputs are the conjunction of the weak constraints, $f$, and the blocking clause $b$ that negates the unique solution to the strong constraints.  The literals of this clause are the negations of the subgoals.

An explanation for the puzzle solution is an explanation of each subgoal.  Hence, the while loop produces an explanation for one of them at each iteration.

In a given iteration, the subgoals are divided into those that, at this stage, admit a weak proof, and those that do not.  if the first class is not empty, the subgoal to be explained is one with a resolution proof of minimum length.  Otherwise, a connectivity argument is required for all remaining subgoals.

The explanation of resolution proofs is delegated to the LLM.  The pictures, in our prototype, are deemed to be self-evident.  This is sometimes not fully justified.  A more refined implementation will ask the user whether the implications that led to the proof by picture need to be further explained.

Once a subgoal has been explained, the unit clause that asserts it is conjoined with $f$.  As a result, some of the remaining subgoals may get shorter weak proofs.  Some other subgoals that had no weak proof may get one.


\section{Additional Hitori Lemmas}
\label{sec:hitori-lemmas}

\begin{figure}
  \centering
  \begin{tikzpicture}[scale=0.5,transform shape]
    \sffamily\sansmath\LARGE
    \colorlet{shaded}{blue}
    \colorlet{background}{yellow!5!white}
    \colorlet{unshaded}{green!20!white}
    \colorlet{assumption}{red}
    \begin{scope}[yshift=5cm]
      \fill[background] (0,0) rectangle (3,3);
      \fill[unshaded] (0,1) rectangle +(1,1);
      \node[black] at (0.5,2.5) {$a$};
      \node[black] at (1.5,2.5) {$a$};
      \draw[step=1cm,ultra thin] (0,0) grid (3,3);
      \draw[very thick] (0,0) |- (3,3);
    \end{scope}
    \begin{scope}[xshift=5cm,yshift=5cm]
      \fill[background] (0,0) rectangle (3,3);
      \fill[unshaded] (1,2) rectangle +(1,1);
      \node[black] at (0.5,1.5) {$a$};
      \node[black] at (1.5,1.5) {$a$};
      \draw[step=1cm,ultra thin] (0,0) grid (3,3);
      \draw[very thick] (0,0) |- (3,3);
    \end{scope}
    \begin{scope}[xshift=10cm,yshift=5cm]
      \fill[background] (0,0) rectangle (6,3);
      \fill[shaded] (4,1) rectangle +(1,1);
      \node[black] at (1.5,1.5) {$a$};
      \node[black] at (2.5,1.5) {$a$};
      \node[background] at (4.5,1.5) {$a$};
      \draw[step=1cm,ultra thin] (0,0) grid (6,3);
    \end{scope}
    \begin{scope}
      \fill[background] (0,0) rectangle (8,4);
      \fill[unshaded] (5,2) rectangle +(1,1);
      \node[black] at (1.5,2.5) {$a$};
      \node[black] at (2.5,2.5) {$b$};
      \node[black] at (4.5,2.5) {$a$};
      \node[black] at (6.5,2.5) {$b$};
      \draw[step=1cm,ultra thin] (0,0) grid (8,4);
    \end{scope}
    \begin{scope}[xshift=10cm]
      \fill[background] (0,0) rectangle (6,4);
      \fill[unshaded] (3,1) rectangle +(1,1);
      \fill[unshaded] (4,2) rectangle +(1,1);
      \node[black] at (1.5,2.5) {$a$};
      \node[black] at (1.5,1.5) {$b$};
      \node[black] at (3.5,2.5) {$a$};
      \node[black] at (4.5,1.5) {$b$};
      \draw[step=1cm,ultra thin] (0,0) grid (6,4);
    \end{scope}
  \end{tikzpicture}
  \caption{A few tricks of the trade.}
  \label{fig:hitori-lemmas}
\end{figure}
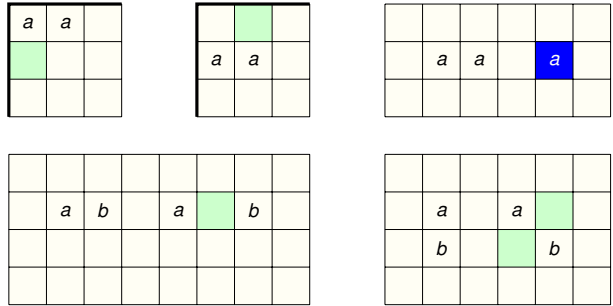

Besides the Sandwich lemma and the Unshaded Neighbor lemma, the
``tricks'' of Figure~\ref{fig:hitori-lemmas} are popular with Hitori
solvers.  In each case, reversing the shading of a cell marked as either shaded or unshaded leads to immediate violation of the rules.
The first two tricks apply to corners of the grid because their inferences come from preventing the trapping of the corner square.  For example, if \texttt{r1c1} and \texttt{r1c2}, in the top-left
corner, contain the same symbol, shading \texttt{r2c1} would cause the
other neighbor of \texttt{r1c1} to be shaded as well, trapping
\texttt{r1c1}.
The others apply everywhere in the grid.  The pattern at the bottom right in Figure~\ref{fig:hitori-lemmas} could be applied to explain why \texttt{r1c2} and \texttt{r2c3} in Section~\ref{sec:complete-p63} of this appendix must be left unshaded (with $a=b=3$).


\section{Prompts}

Prompt used for ``partial-state'' examples can be found in Figure \ref{fig:prompt}.
Recall that in ``partial-state'' examples, we only show history relevant to the task at hand. 
In order to get the prompt used for ``full-state'', we  mention all the states prior, regardless of it finds use in solving the local step.

\begin{figure}[h]
\small

\begin{lstlisting}[basicstyle=\ttfamily, 
                frame=single,
                keywordstyle=,
                commentstyle=,
                stringstyle=,
                ]
Look at the following step taken from a Z3 proof of a Hitori puzzle's solution. 

```text 
    (unit-resolution (asserted (or c01_05 c02_05)) (asserted (not c01_05)) c02_05)
```

We obtain the meaning of the clauses as follows: 

```text
Reasons for the clauses in the proof
    (or c01_05 c02_05): r1c5 and r2c5 cannot both be shaded because they share an edge
```

Before this step, the cell `r1c5` was shaded.

After this step, the cell `r2c5` was left unshaded. 

You could explain this to a human in the following way: 

```text
r1c5 and r2c5 share an edge. Hence, they cannot both be shaded. Since r1c5 is already shaded, r2c5 is left unshaded. 
```

Similarly, we can look at another step taken from a Z3 proof of a Hitori puzzle's solution. 

```text
(let ((a!1 (unit-resolution (asserted (or (not c01_07) (not c07_07)))
    (asserted c07_07)
    (not c01_07))))
    (unit-resolution a!1 (asserted c01_07) false))
```

We obtain the meaning of the clauses as follows: 

```text 
Reasons for the clauses in the proof
    c07_07: r7c7 has identical neighbors in its column
    (or (not c01_07) (not c07_07)): r1c7 and r7c7 cannot both be unshaded because they have the same symbol
```


 

\end{lstlisting}

\caption{``Partial-State'' Prompt - Part $1$}
\label{fig:prompt}
\end{figure}

\begin{figure}[h]
\small
    \ContinuedFloat

    \begin{lstlisting}[basicstyle=\ttfamily, 
                frame=single,
                keywordstyle=,
                commentstyle=,
                stringstyle=,
                ]

Before this step, the cell `r7c7` was unshaded

After this step, the cell `r1c7` was shaded. 

You could explan this to a human in the following way: 

```text 
r1c7 and r7c7 are in the same column, and contain the same symbol. Since r7c7 is unshaded, r1c7 must be shaded. 
Asserting that r1c7 is unshaded leads to a contradiction. Hence, r1c7 has to be shaded. 
```

    
Here is another example step taken from a Z3 proof of a Hitori puzzle's solution: 

```text
(unit-resolution (unit-resolution (asserted (or c04_03 c05_02 c05_04 c06_03))
    (asserted (not c04_03))
    (asserted (not c05_02))
    (asserted (not c05_04))
    c06_03)
    (asserted (not c06_03))
false)
```

We obtain the meaning of the clauses as follows: 

```text 
Reasons for the clauses in the proof
(or c04_03 c05_02 c05_04 c06_03): r5c3 must have an unshaded neighbor
```

Before this step, the cells `r4c3`, `r5c2`, `r5c4` were unshaded.

After this step, the cell `r6c3` was left unshaded. 

You could explain this to a human in the following way: 

```text
Since the other 3 neighbors of r5c3 are already shaded, if we shaded r6c3, r5c3 would be trapped by shaded cells. 
This would prevent r5c3 from being connected to other cells as per Hitori rules. 
Hence r6c3 must be left unshaded. 
```


    
\end{lstlisting}

\caption{``Partial-State'' Prompt - Part $2$}
\label{fig:prompt2}
\end{figure}

\begin{figure}[h]
\small
    \ContinuedFloat

    \begin{lstlisting}[basicstyle=\ttfamily, 
                frame=single,
                keywordstyle=,
                commentstyle=,
                stringstyle=,
                ]


    
Now, look at the following step taken from a Z3 proof of a Hitori puzzle's solution. 

```text
{z3}
```

Remember, `ci_j` means cell[i,j] must be unshaded. 
Conversely, `not ci_j` means cell[i,j] must be shaded. 

We obtain the meaning of the clauses as follows:

```text
{solver_explanation}
```

{shaded_cells_message}
{unshaded_cells_message}

After this step, the cell `{target}` was {value}.

Prepare an explanation for this step that would be understandable to a human.

        

    \end{lstlisting}
    
    \caption{``Partial-State'' Prompt - Part $3$}
    \label{fig:prompt3}
\end{figure}


\section{Annotation Results}

The authors served as annotators for assessing the LLM's translation of $Z3$ proofs into natural language. 
In this section, we disclose additional statistics gathered from our annotations. 
In order to conduct this study, we collapse our Likert scale into just $3$ divisions - \texttt{agree, neutral, disagree}. 

\subsection{Annotator Agreements}

Outlined in Table \ref{tab:anno-agreement}.

\begin{table}[h]
\centering
\caption{Cohen's Kappa ($\kappa$) agreement reported on annotator's assessment of the LLM responses' correctness.}
\begin{tabular}{@{}l|l|l@{}}
\toprule
              & Annotator Pair 1 (Trivial) & Annotator Pair 2 (Complex) \\ \midrule
Full-State    & 0.55             & 0.63             \\
Partial-State & 0.27             & 0.57             \\ \midrule
Overall       & 0.42             & 0.60            
\end{tabular}
\label{tab:anno-agreement}
\end{table}

\subsection{Descriptive Statistics of Annotations}

In this section, we share average annotator results across multiple dimensions, for ``full'', ``partial'' and combined states - as provided in prompts (Tables \ref{tab:full_partial}, \ref{tab:partial}, \ref{tab:full}). Our scale goes from $1$ (Strongly Disagree) to $5$ (Strongly Agree).
Generally, higher scores indicate better LLM generation, except for the dimension ``Falsities'', where the lower score is better. 

\begin{table}[h]
\centering
\caption{Average ratings as assigned by annotator pairs, across different dimensions. These are ratings for the \textbf{``full-state'' and ``partial-state''} settings combined. }\label{tab:full_partial}
\begin{tabular}{@{}c|cc|cc@{}}
\toprule
\multirow{2}{*}{Dimension} & \multicolumn{2}{c|}{Annotator Pair 1 (Trivial)} & \multicolumn{2}{c}{Annotator Pair 2 (Complex)} \\ \cmidrule(l){2-5} 
                           & \multicolumn{1}{c|}{Annotator 1}  & Annotator 2 & \multicolumn{1}{c|}{Annotator 1} & Annotator 2 \\ \midrule
Correctness                & 2.6                               & 3.56        & 4.52                             & 4.36        \\
Relevance                  & 3.00                              & 4.88        & 4.92                             & 4.28        \\
Completeness               & 3.48                              & 4.88        & 4.72                             & 4.28        \\
Falsities                  & 3.40                              & 1.68        & 1.56                             & 1.60        \\
Clarity                    & 3.88                              & 4.80        & 4.36                             & 4.20        \\
Consciseness               & 3.84                              & 4,28        & 4.56                             & 1.96     \\ \bottomrule 
\end{tabular}
\end{table}

\begin{table}[h]
\centering
\caption{Average ratings as assigned by annotator pairs, across different dimensions. These are ratings for \textbf{``partial-state''} settings combined. }\label{tab:partial}
\begin{tabular}{@{}c|cc|cc@{}}
\toprule
\multirow{2}{*}{Dimension} & \multicolumn{2}{c|}{Annotator Pair 1 (Trivial)} & \multicolumn{2}{c}{Annotator Pair 2 (Complex)} \\ \cmidrule(l){2-5} 
                           & \multicolumn{1}{c|}{Annotator 1}  & Annotator 2 & \multicolumn{1}{c|}{Annotator 1} & Annotator 2 \\ \midrule
Correctness                & 2.76                              & 3.72        & 4.60                             & 4.20        \\
Relevance                  & 2.92                              & 4.92        & 4.92                             & 3.88        \\
Completeness               & 2.92                              & 5.00        & 4.68                             & 4.36        \\
Falsities                  & 3.24                              & 1.72     & 1.48                             & 1.88        \\
Clarity                    & 3.88                              & 4.76        & 4.60                             & 4.04        \\
Consciseness               & 3.88                              & 4.60        & 4.76                             & 1.80        \\ \bottomrule
\end{tabular}
\end{table}

\begin{table}[h]
\centering
\caption{Average ratings as assigned by annotator pairs, across different dimensions. These are ratings for \textbf{``full-state''} settings combined. }\label{tab:full}
\begin{tabular}{@{}c|cc|cc@{}}
\toprule
\multirow{2}{*}{Dimension} & \multicolumn{2}{c|}{Annotator Pair 1 (Trivial)} & \multicolumn{2}{c}{Annotator Pair 2 (Complex)} \\ \cmidrule(l){2-5} 
                           & \multicolumn{1}{c|}{Annotator 1}  & Annotator 2 & \multicolumn{1}{c|}{Annotator 1} & Annotator 2 \\ \midrule
Correctness                & 2.44                              & 3.40        & 4.44                             & 4.52        \\
Relevance                  & 3.08                              & 4.84        & 4.92                             & 4.68        \\
Completeness               & 4.04                              & 4.76        & 4.76                             & 4.20        \\
Falsities                  & 3.56                              & 1.64        & 1.64                             & 1.32        \\
Clarity                    & 3.88                              & 4.84        & 4.12                             & 4.36        \\
Consciseness               & 3.80                              & 3.96        & 4.36                             & 2.12        \\ \bottomrule
\end{tabular}
\end{table}
\newpage

\section{Complete explanation of the Hitori puzzle from Figure~\ref{fig:hitori-p63}}
\label{sec:complete-p63}
To illustrate the potential of our approach in explaining Hitori puzzles, consider the $5{\times}5$ puzzle shown in Figure~\ref{fig:hitori-p63}, along with its solution. Since this puzzle has a unique solution, asking Z3 to prove its uniqueness yields explanations for the shading and unshading decisions. However, the corresponding resolution proof is \num{82438} characters long, making it infeasible to understand either manually or directly via an LLM. Consequently, users of our tool begin by interactively querying the first logical step in the puzzle.

A commonly useful pattern in a Hitori grid is the \emph{sandwich}, where a symbol is surrounded---either horizontally or vertically---by two identical symbols. One of these two identical symbols must be shaded, which forces the sandwiched symbol to remain unshaded due to adjacency constraints. 
In our example, there are five sandwiched squares (notably, \texttt{r3c4} is sandwiched both horizontally and vertically). We begin by examining the tool-generated explanation for \texttt{r1c4}, a representative sandwich case. The explanations in this section were generated using ChatGPT-4o.

\begin{mybox}{\texttt{r1c4}}
\footnotesize{\textbf{Z3 Proof} for \texttt{r1c4} unshaded:}
 \begin{lstlisting}[basicstyle=\scriptsize\ttfamily]
c01_04
\end{lstlisting}

\footnotesize{\textbf{LLM Explanation:} This cell lies between two identical numbers (3s) in its row. According to the Sandwich Lemma, if two identical numbers are separated by exactly one cell, that middle cell must be unshaded. Shading it would allow both 3s on either side to remain unshaded, violating the Uniqueness Rule. Therefore, (1,4) must be unshaded.}

\end{mybox}

\begin{figure}[t]
  \centering
  \begin{minipage}[b]{0.32\textwidth}
    \centering
    \includegraphics[width=\textwidth]{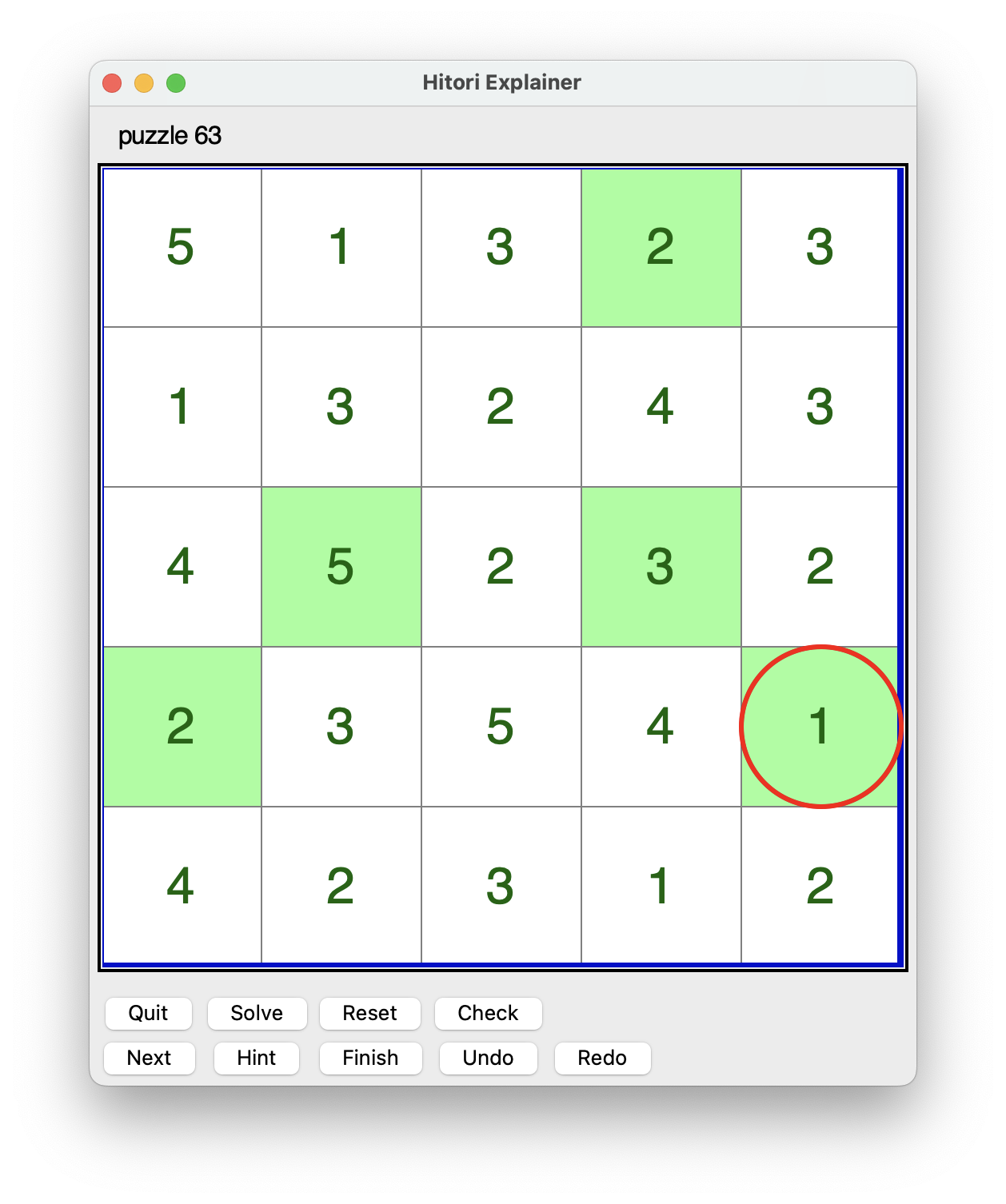}
    \caption*{(a) Deciding \texttt{r1c4}---\texttt{r4c5}.}
  \end{minipage}
  \hfill
  \begin{minipage}[b]{0.32\textwidth}
    \centering
    \includegraphics[width=\textwidth]{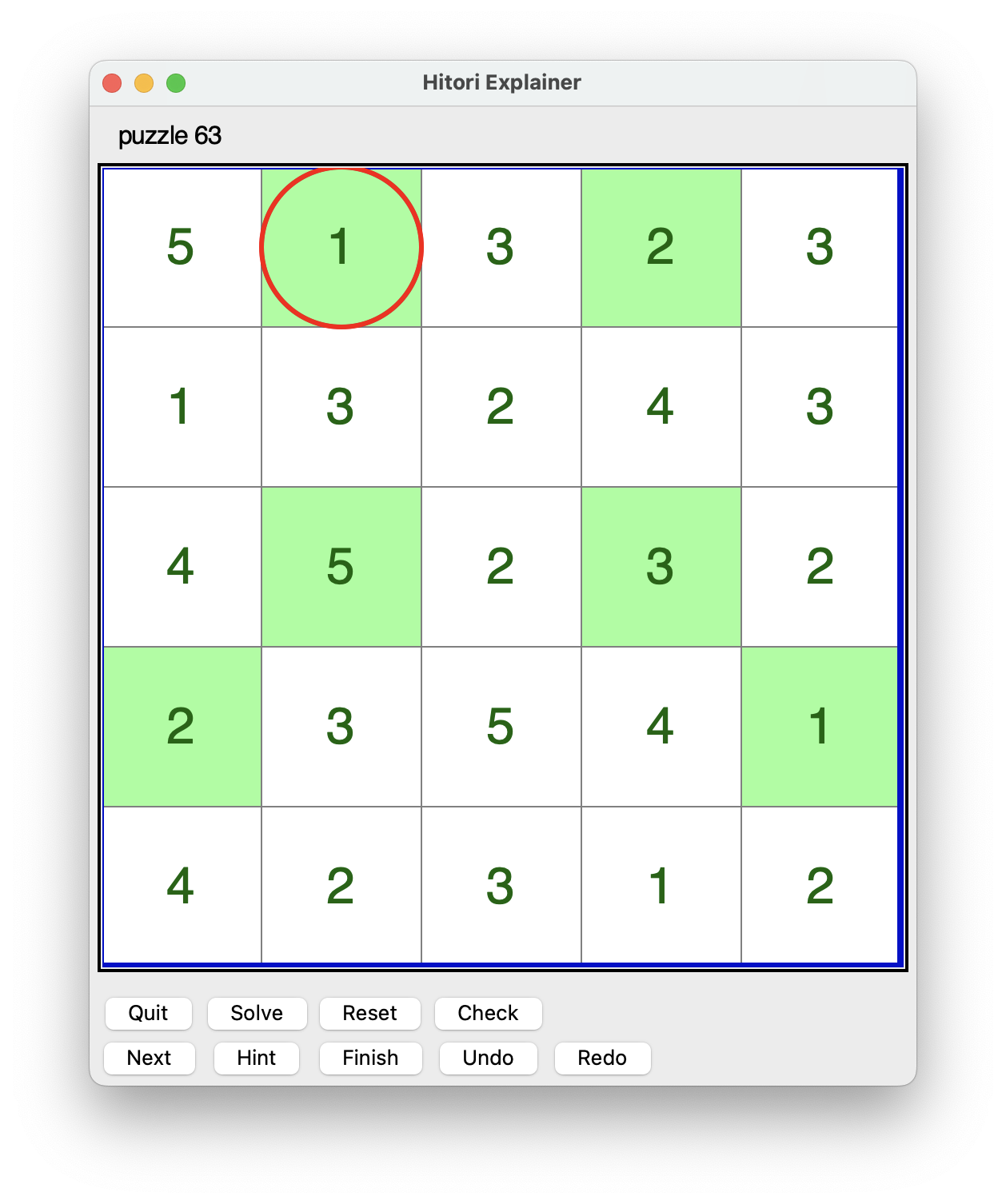}
    \caption*{(b) Deciding Cell \texttt{r1c2}}
  \end{minipage}
  \hfill
  \begin{minipage}[b]{0.32\textwidth}
    \centering
    \includegraphics[width=\textwidth]{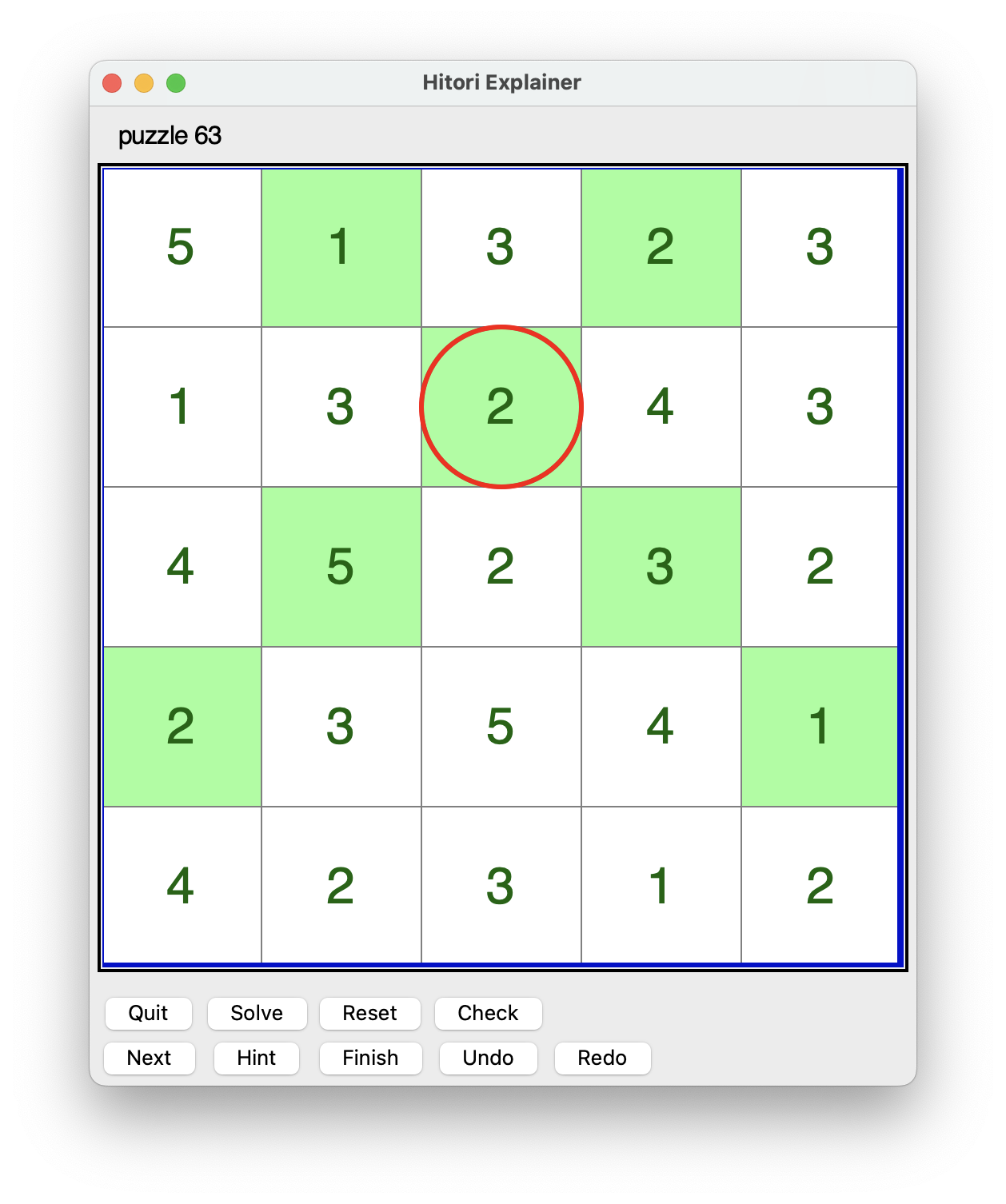}
    \caption*{(c) Deciding Cell \texttt{r2c3}}
  \end{minipage}
  \caption{Stepwise justification of key shading and unshading decisions in the Hitori puzzle (1).}
  \label{fig:three-images}
\end{figure}

The explanations for other sandwiched cells---\texttt{r3c2}, \texttt{r3c4}, and \texttt{r4c5}---follow similar reasoning and are omitted for brevity. Their positions are highlighted in Figure~\ref{fig:three-images}(a).

We now move on to reasoning that involves longer chains of logical consequences rather than local sandwich structures. The next two cases, \texttt{r1c2} (Figure~\ref{fig:three-images}(b)) and \texttt{r2c3} (Figure~\ref{fig:three-images}(c)), illustrate how contradictions emerge from assuming a cell is shaded.

\begin{mybox}{\texttt{r1c2}}
\footnotesize{\textbf{Z3 Proof} for \texttt{r1c2} unshaded:}
    \begin{lstlisting}[basicstyle=\scriptsize\ttfamily]
(let ((a!1 (unit-resolution (asserted (or (not c02_02) (not c02_05)))
          (unit-resolution (asserted (or c01_02 c02_02))
                          (asserted (not c01_02))
                          c02_02)
          (not c02_05)))
      (a!2 (unit-resolution (asserted (or (not c01_03) (not c01_05)))
          (unit-resolution (asserted (or c01_02 c01_03))
                          (asserted (not c01_02))
                          c01_03)
          (not c01_05))))
  (unit-resolution (asserted (or c01_05 c02_05)) a!1 a!2 false))
\end{lstlisting}

\footnotesize{\textbf{LLM Explanation:}  
This cell must be \emph{unshaded}. If \texttt{r1c2} were shaded, then:
\begin{itemize}
    \item \texttt{r1c3} would have to be unshaded to satisfy the Separation Rule.
    \item But \texttt{r1c3} and \texttt{r1c5} have the same symbol (3), so both cannot remain unshaded. This forces \texttt{r1c5} to be shaded.
    \item Similarly, shading \texttt{r1c2} forces \texttt{r2c2} to be unshaded due to adjacency.
    \item Now, \texttt{r2c2} and \texttt{r2c5} have the same symbol (3), so one must be shaded. Since \texttt{r1c5} is already shaded, \texttt{r2c5} must remain unshaded.
    \item This leads to \texttt{r1c5} and \texttt{r2c5} both being shaded and adjacent, violating the Separation Rule.
\end{itemize}}

Thus, assuming \texttt{r1c2} is shaded leads to a contradiction. Therefore, it must be unshaded.

\end{mybox}

\begin{mybox}{\texttt{r2c3}}
\footnotesize{\textbf{Z3 Proof} for \texttt{r2c3} unshaded:}
    \begin{lstlisting}[basicstyle=\scriptsize\ttfamily]
(let ((a!1 (unit-resolution (asserted (or (not c01_03) (not c01_05)))
                            (unit-resolution (asserted (or c01_03 c02_03))
                                             (asserted (not c02_03))
                                             c01_03)
                            (not c01_05)))
      (a!2 (unit-resolution (asserted (or (not c02_02) (not c02_05)))
                            (unit-resolution (asserted (or c02_02 c02_03))
                                             (asserted (not c02_03))
                                             c02_02)
                            (not c02_05))))
  (unit-resolution (asserted (or c01_05 c02_05)) a!1 a!2 false))
\end{lstlisting}

\footnotesize{\textbf{LLM Explanation:}  
This cell must be \emph{unshaded}. If we assume \texttt{r2c3} is shaded, then:
\begin{itemize}
    \item \texttt{r2c2} must be unshaded (due to adjacency).
    \item But \texttt{r2c2} and \texttt{r2c5} contain the same symbol (3), so they can’t both be unshaded — thus, \texttt{r2c5} must be shaded.
    \item Also, shading \texttt{r2c3} forces \texttt{r1c3} to be unshaded (adjacency).
    \item But \texttt{r1c3} and \texttt{r1c5} also contain the same symbol (3), so \texttt{r1c5} must be shaded.
    \item Now both \texttt{r1c5} and \texttt{r2c5} are shaded, but they’re adjacent — violating the Separation Rule.
\end{itemize}}

Thus, assuming \texttt{r2c3} is shaded leads to a contradiction. Therefore, it must be unshaded.

\end{mybox}

Figure~\ref{fig:three-images}(b,c) captures these key decisions.

The sequence continues with a critical inference involving vertical duplication: if \texttt{r2c3} is unshaded, then \texttt{r3c3} must be shaded. This decision cascades into further adjacency constraints on \texttt{r4c3} and has wider implications for the puzzle.

\begin{figure}[t]
  \centering
  \begin{minipage}[b]{0.32\textwidth}
    \centering
    \includegraphics[width=\textwidth]{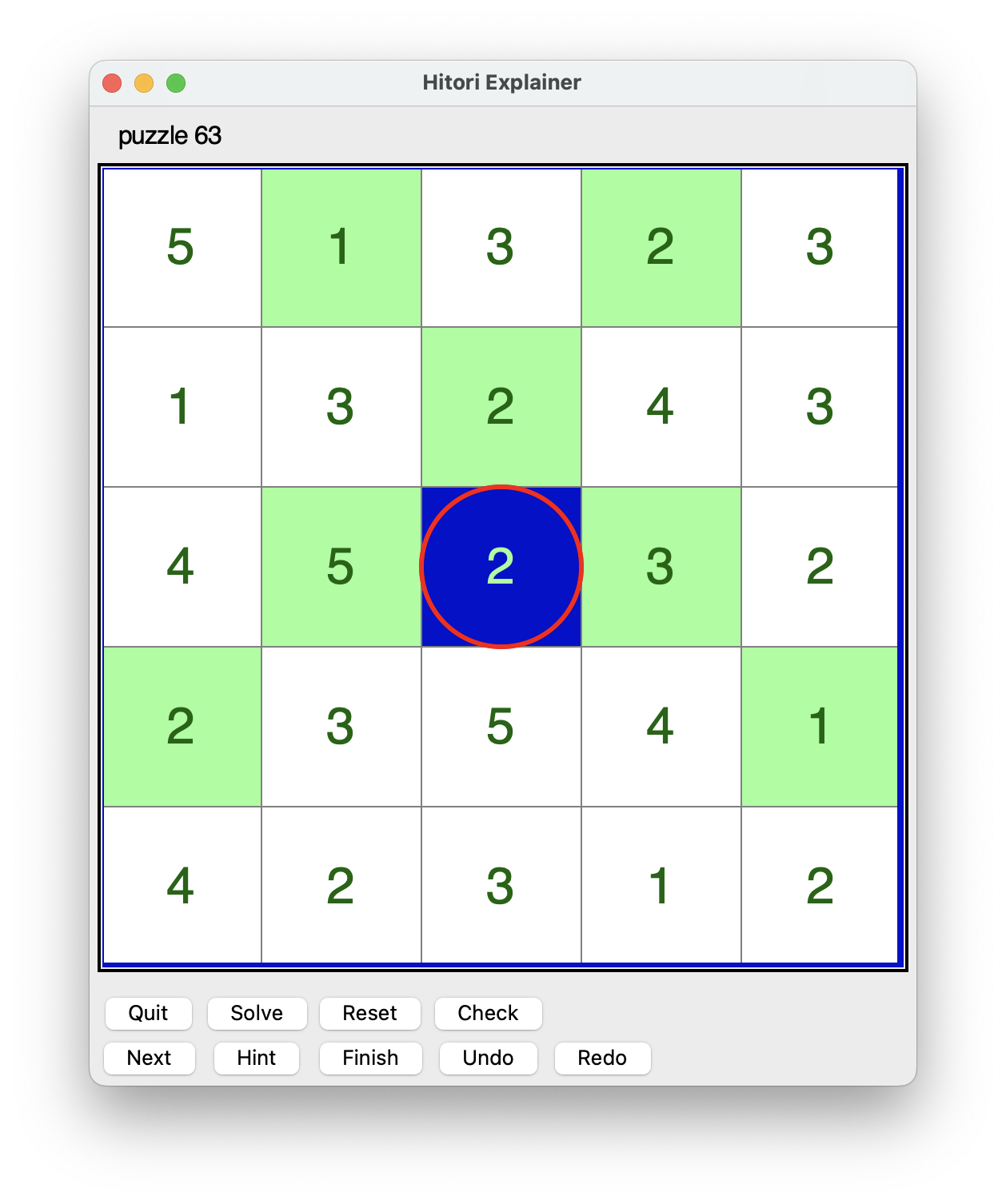}
    \caption*{(a) Deciding \texttt{r3c3}.}
  \end{minipage}
  \hfill
  \begin{minipage}[b]{0.32\textwidth}
    \centering
    \includegraphics[width=\textwidth]{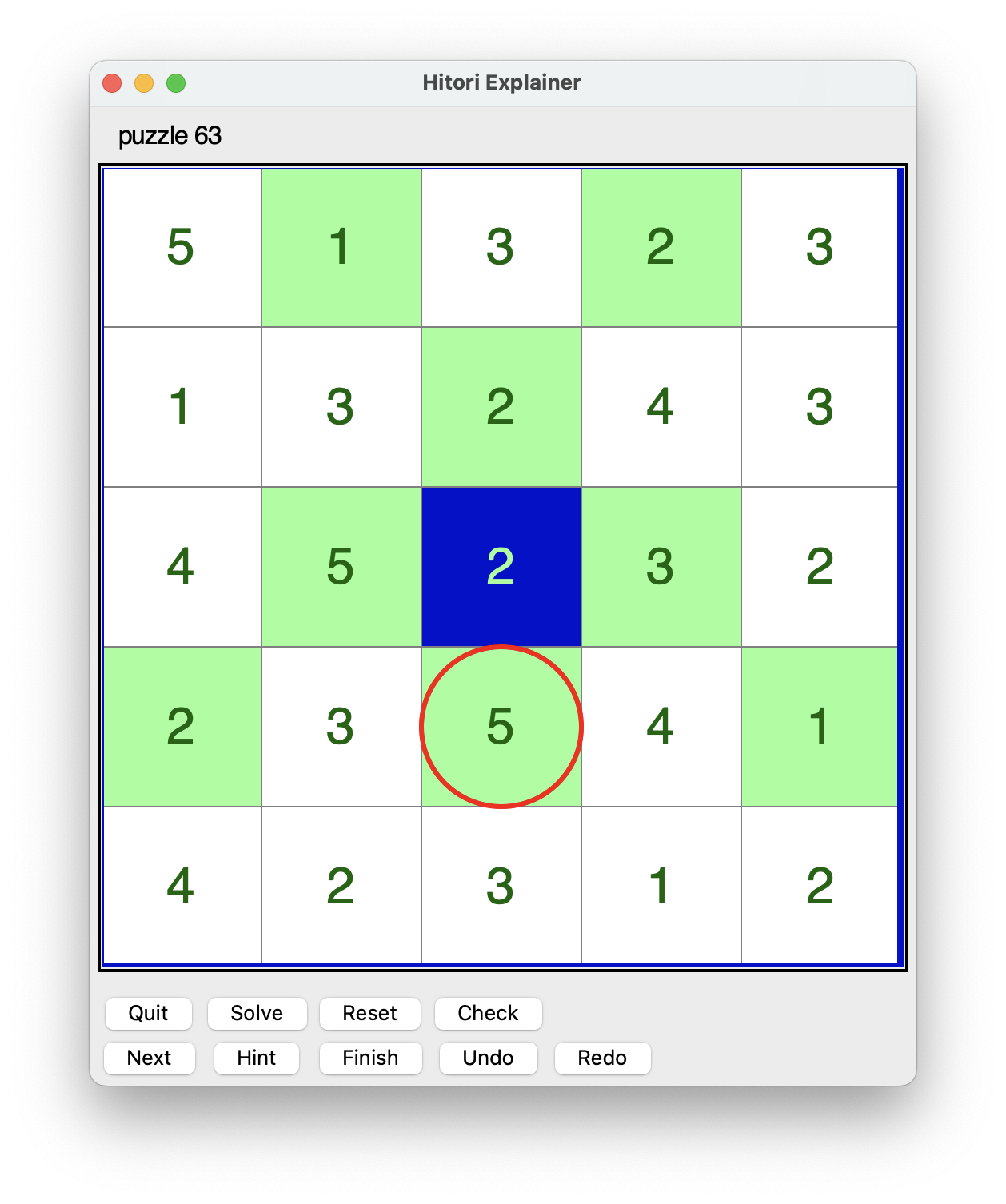}
    \caption*{(b) Deciding Cell \texttt{r4c3}}
  \end{minipage}
  \hfill
  \begin{minipage}[b]{0.32\textwidth}
    \centering
    \includegraphics[width=\textwidth]{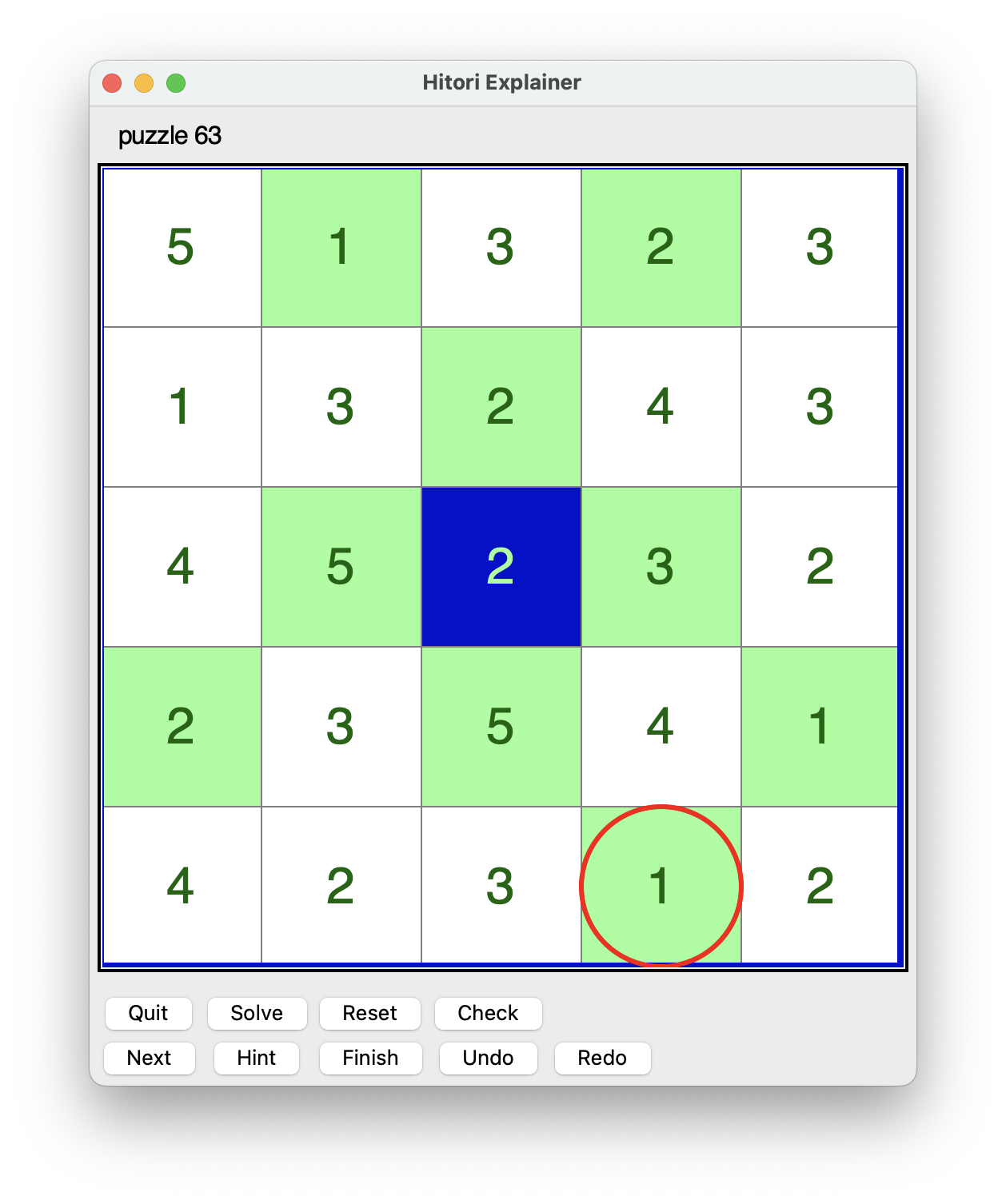}
    \caption*{(c) Deciding Cell \texttt{r5c4}}
  \end{minipage}
  \caption{Stepwise justification of key shading and unshading decisions in the Hitori puzzle (2).}
  \label{fig:three-images-two}
\end{figure}

\begin{mybox}{\texttt{r3c3}}
\footnotesize{\textbf{Z3 Proof} for \texttt{r3c3} shaded:}
    \begin{lstlisting}[basicstyle=\scriptsize\ttfamily]
(let ((a!1 (unit-resolution (asserted (or (not c02_03) (not c03_03)))
                            (asserted c02_03)
                            (not c03_03))))
  (unit-resolution a!1 (asserted c03_03) false))
    \end{lstlisting}

\footnotesize{\textbf{LLM Explanation:}  
Cell (3,3): \emph{Shaded}  
Reason: This cell contains the same symbol as (2,3), which has already been established as unshaded. According to the Uniqueness Rule, two cells in the same column with the same symbol cannot both be unshaded. So (3,3) must be shaded.}
\end{mybox}

\begin{mybox}{\texttt{r4c3}}
\footnotesize{\textbf{Z3 Proof} for \texttt{r4c3} unshaded:}
    \begin{lstlisting}[basicstyle=\scriptsize\ttfamily]
(unit-resolution (asserted (or c03_03 c04_03)) (asserted (not c03_03)) c04_03)
\end{lstlisting}

\footnotesize{\textbf{LLM Explanation:}  
Cell (4,3): \emph{Unshaded}  
Reason: Since (3,3) is already known to be shaded, and (4,3) shares an edge with it, the two cannot both be shaded — that would violate the Separation Rule. Therefore, (4,3) must be unshaded.}
\end{mybox}

Next, we consider the case of \texttt{r5c4}, where a longer chain of dependencies must be resolved to avoid violating both the Separation Rule and the Unshaded Neighbor Lemma.

\begin{mybox}{\texttt{r5c4}}
\footnotesize{\textbf{Z3 Proof} for \texttt{r5c4} unshaded:}
    \begin{lstlisting}[basicstyle=\scriptsize\ttfamily]
(let ((a!1 (unit-resolution (asserted (or (not c02_04) (not c04_04)))
                            (unit-resolution (asserted (or c04_04 c05_04))
                                             (asserted (not c05_04))
                                             c04_04)
                            (not c02_04)))
      (a!2 (unit-resolution (asserted (or (not c01_03) (not c05_03)))
                            (unit-resolution (asserted (or c05_03 c05_04))
                                             (asserted (not c05_04))
                                             c05_03)
                            (not c01_03))))
  (unit-resolution (asserted (or (not c01_05) (not c02_05)))
                   (unit-resolution (asserted (or c02_04 c02_05)) a!1 c02_05)
                   (unit-resolution (asserted (or c01_03 c01_05 c02_04))
                                    a!2
                                    a!1
                                    c01_05)
                   false))
    \end{lstlisting}

\footnotesize{\textbf{LLM Explanation:}  
Cell (5,4): \emph{Unshaded}  
Reason:  
Let’s walk through the reasoning step by step. This proof rules out the possibility of shading cell (5,4), because doing so would cause a contradiction involving several interconnected cells and rules:

\begin{itemize}
    \item \textbf{Step 1:} If (5,4) were shaded...  
    (5,3) must be unshaded (adjacent cells cannot both be shaded — Separation Rule).  
    But (5,3) and (1,3) contain the same number (3), so both cannot be unshaded (Uniqueness Rule, column). Hence, (1,3) must be shaded.

    \item \textbf{Step 2:} Still assuming (5,4) is shaded...  
    (4,4) must be unshaded (shares an edge with (5,4)).  
    But (4,4) and (2,4) both contain 4, and so cannot both be unshaded — one of them must be shaded.  
    Since we’re assuming (4,4) is unshaded, this forces (2,4) to be shaded.

    \item \textbf{Step 3:} Now with (2,4) shaded...  
    Then (2,5) must be unshaded (adjacent to shaded (2,4)).  
    But (2,5) and (1,5) are in the same column and contain the same number (3), so both can’t be unshaded.  
    To fix this, we must shade (1,5).

    \item \textbf{Step 4:} Now, with (1,5) shaded and (1,3) shaded (from earlier), consider (1,4)...  
    Cell (1,4) must have at least one unshaded neighbor (due to the Unshaded Neighbor Lemma).  
    But its neighbors are (1,3), (1,5), and (2,4) — all of which we've forced to be shaded.  
    This violates the Unshaded Neighbor Lemma, creating a contradiction.
\end{itemize}

\textbf{Conclusion:}  
Assuming (5,4) is shaded leads to a cascade of forced decisions that eventually violate the Unshaded Neighbor Lemma at (1,4).  
Thus, (5,4) must be unshaded.}
\end{mybox}
The cell \texttt{r2c1} demonstrates how assumptions about shading can violate the connectivity of unshaded cells---a key constraint in Hitori.

\begin{mybox}{\texttt{r2c1}}
\footnotesize{\textbf{Z3 Proof} for \texttt{r2c1} unshaded:}
    \begin{lstlisting}[basicstyle=\scriptsize\ttfamily]
Step 11. Connectedness lost if r2c1 is shaded
ASCII grid:
. . x . .
x . . . x
. . X . .
. x - x .
x - - - x
    \end{lstlisting}
    
\footnotesize{\textbf{LLM Explanation:}  
Cell (2,1): \emph{Unshaded}  
Reason: If \texttt{(2,1)} were shaded, it would disconnect the unshaded cells into two separate groups — violating the Connectivity Rule.

The ASCII grid shows this clearly:
\begin{itemize}
    \item Cells marked with \texttt{.} form one connected component of unshaded cells.
    \item Cells marked with \texttt{-} form another, unreachable from the first if \texttt{(2,1)} is shaded.
    \item Cells marked with \texttt{x} are shaded as a result of the assumption that \texttt{(2,1)} is shaded.
\end{itemize}

Since unshaded cells must remain part of a single connected area, shading \texttt{(2,1)} is not allowed.  
Therefore, \texttt{(2,1)} must be unshaded.}
\end{mybox}

The logical consequences of the above steps affect several additional cells. We briefly note that the proofs for \texttt{r2c2}, \texttt{r1c1}, and \texttt{r1c3} rely on similar connectivity arguments and are omitted here.

We now turn to the region involving \texttt{r1c5} and its adjacent cells, showing how shading propagates under the Uniqueness and Separation constraints.

\begin{mybox}{\texttt{r1c5}}
\footnotesize{\textbf{Z3 Proof} for \texttt{r1c5} shaded:}
    \begin{lstlisting}[basicstyle=\scriptsize\ttfamily]
(let ((a!1 (unit-resolution (asserted (or (not c01_03) (not c01_05)))
                            (asserted c01_03)
                            (not c01_05))))
  (unit-resolution a!1 (asserted c01_05) false))
    \end{lstlisting}

\footnotesize{\textbf{LLM Explanation:}  
Cell (1,5): \emph{Shaded}  
Reason:  
Cells \texttt{(1,3)} and \texttt{(1,5)} are in the same row and contain the same number (3).  
Since \texttt{(1,3)} is already known to be unshaded, \texttt{(1,5)} must be shaded to avoid violating the Uniqueness Rule.}
\end{mybox}

\begin{mybox}{\texttt{r2c5}}
\footnotesize{\textbf{Z3 Proof} for \texttt{r2c5} unshaded:}
    \begin{lstlisting}[basicstyle=\scriptsize\ttfamily]
(unit-resolution (asserted (or c01_05 c02_05)) (asserted (not c01_05)) c02_05)
    \end{lstlisting}

\footnotesize{\textbf{LLM Explanation:}  
Cell (2,5): \emph{Unshaded}  
Reason:  
\texttt{(1,5)} is already shaded, and \texttt{(2,5)} shares an edge with it.  
According to the Separation Rule, adjacent cells cannot both be shaded.  
Therefore, \texttt{(2,5)} must be unshaded.}
\end{mybox}

\begin{figure}[t]
  \centering
  \begin{minipage}[b]{0.32\textwidth}
    \centering
    \includegraphics[width=\textwidth]{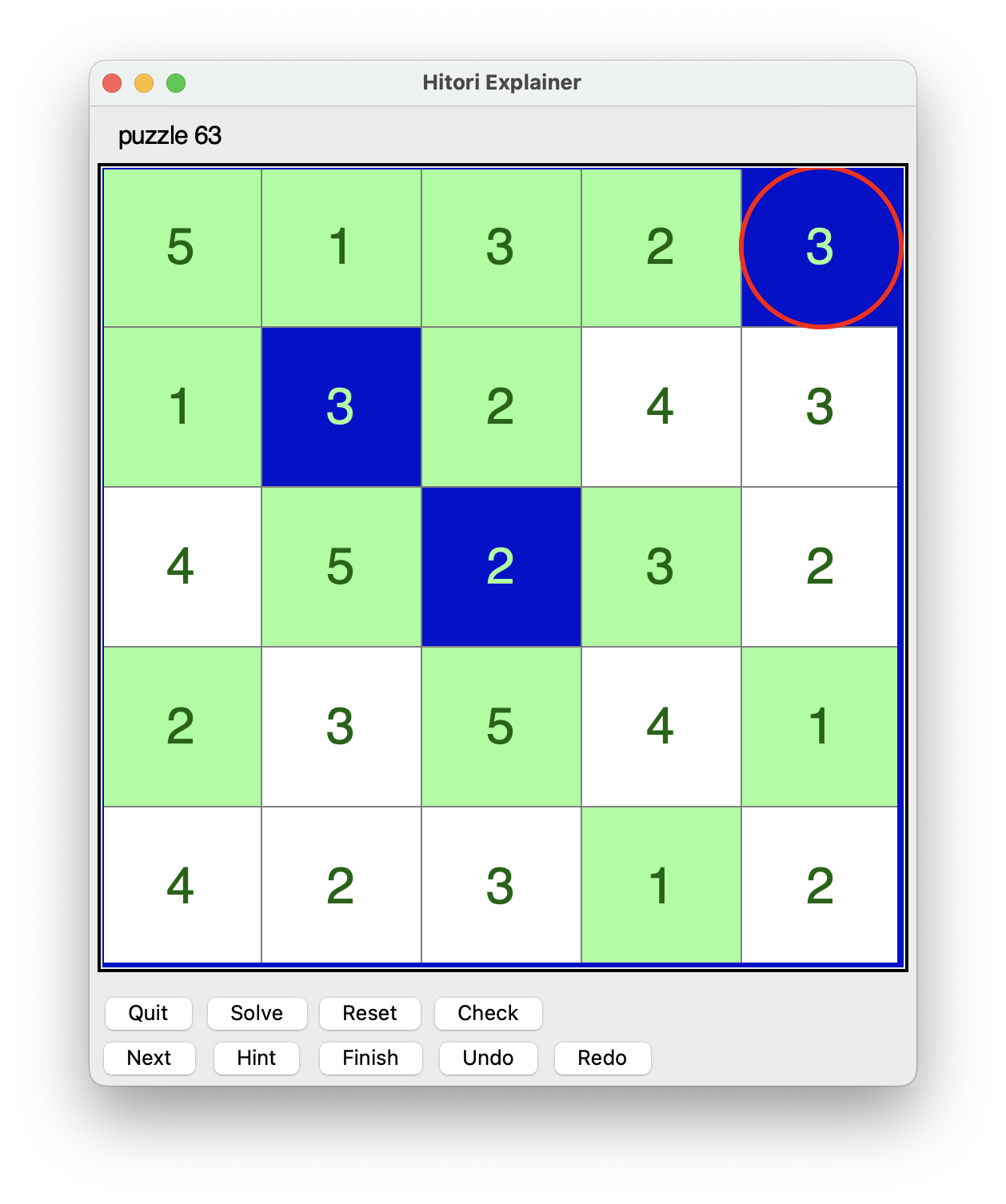}
    \caption*{(a) Deciding \texttt{r1c5}.}
  \end{minipage}
  \hfill
  \begin{minipage}[b]{0.32\textwidth}
    \centering
    \includegraphics[width=\textwidth]{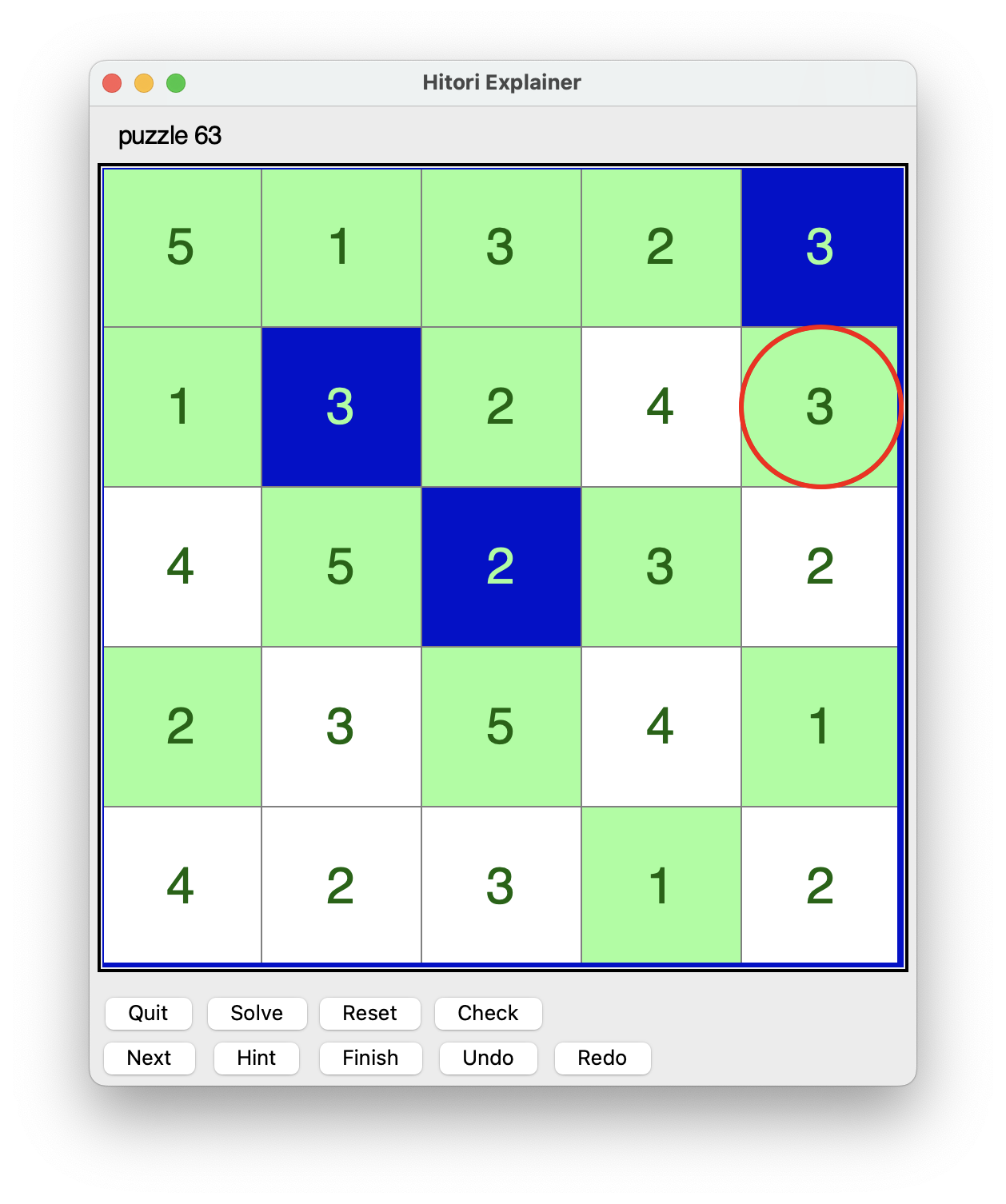}
    \caption*{(b) Deciding Cell \texttt{r2c5}}
  \end{minipage}
  \hfill
  \begin{minipage}[b]{0.32\textwidth}
    \centering
    \includegraphics[width=\textwidth]{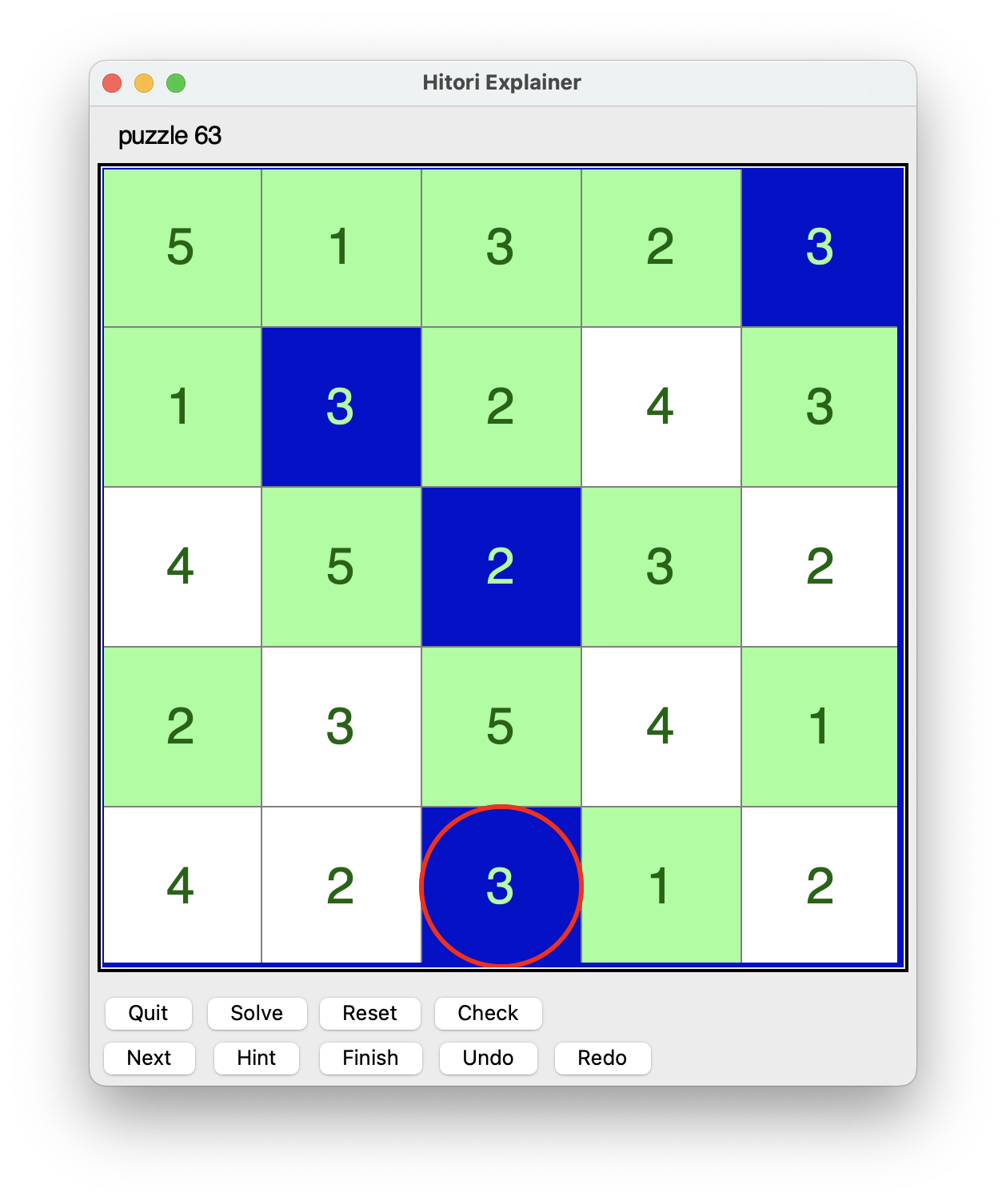}
    \caption*{(c) Deciding Cell \texttt{r5c3}}
  \end{minipage}
   \caption{Stepwise justification of key shading and unshading decisions in the Hitori puzzle (3).}
  \label{fig:three-images-three}
\end{figure}

\begin{mybox}{\texttt{r5c3}}
\footnotesize{\textbf{Z3 Proof} for \texttt{r5c3} shaded:}
    \begin{lstlisting}[basicstyle=\scriptsize\ttfamily]
(let ((a!1 (unit-resolution (asserted (or (not c01_03) (not c05_03)))
                            (asserted c01_03)
                            (not c05_03))))
  (unit-resolution a!1 (asserted c05_03) false))
    \end{lstlisting}

\footnotesize{\textbf{LLM Explanation:}  
Cell (5,3): \emph{Shaded}  
Reason:  
Cells \texttt{(1,3)} and \texttt{(5,3)} are in the same column and contain the same number (3).  
Since \texttt{(1,3)} is unshaded, \texttt{(5,3)} must be shaded to preserve the Uniqueness Rule.}
\end{mybox}

This implies further decisions on cells adjacent to \texttt{r5c3}, as shown below.

\begin{mybox}{\texttt{r5c2}}
\footnotesize{\textbf{Z3 Proof} for \texttt{r5c2} unshaded:}
    \begin{lstlisting}[basicstyle=\scriptsize\ttfamily]
(unit-resolution (asserted (or c05_02 c05_03)) (asserted (not c05_03)) c05_02)
    \end{lstlisting}

\footnotesize{\textbf{LLM Explanation:}  
Cell (5,2): \emph{Unshaded}  
Reason:  
\texttt{(5,3)} is known to be shaded, and \texttt{(5,2)} is adjacent to it.  
To avoid two shaded cells sharing an edge, \texttt{(5,2)} must be unshaded.}
\end{mybox}

\begin{mybox}{\texttt{r5c5}}
\footnotesize{\textbf{Z3 Proof} for \texttt{r5c5} shaded:}
    \begin{lstlisting}[basicstyle=\scriptsize\ttfamily]
(let ((a!1 (unit-resolution (asserted (or (not c05_02) (not c05_05)))
                            (asserted c05_02)
                            (not c05_05))))
  (unit-resolution a!1 (asserted c05_05) false))
    \end{lstlisting}

\footnotesize{\textbf{LLM Explanation:}  
Cell (5,5): \emph{Shaded}  
Reason:  
Cells \texttt{(5,2)} and \texttt{(5,5)} are in the same row and both contain the number 2.  
Since \texttt{(5,2)} has been determined to be unshaded, \texttt{(5,5)} must be shaded to comply with the Uniqueness Rule.}
\end{mybox}
As this region stabilizes, we return to the consequences for \texttt{r4c4} and the subsequent shading of \texttt{r2c4}.

\begin{mybox}{\texttt{r4c4}}
\footnotesize{\textbf{Z3 Proof} for \texttt{r4c4} unshaded:}
    \begin{lstlisting}[basicstyle=\scriptsize\ttfamily]
(unit-resolution (unit-resolution (asserted (or c04_04 c05_03 c05_05))
                                  (asserted (not c05_03))
                                  (asserted (not c05_05))
                                  c04_04)
                 (asserted (not c04_04)) false)
    \end{lstlisting}

\footnotesize{\textbf{LLM Explanation:}  
Cell (4,4): \emph{Unshaded}  
Reason:  
Cell \texttt{(5,4)} must have at least one unshaded neighbor (by the Unshaded Neighbor Lemma).  
However, \texttt{(5,3)} and \texttt{(5,5)} are both shaded, leaving only \texttt{(4,4)} as a potential neighbor.  
Therefore, \texttt{(4,4)} must be unshaded to satisfy this requirement.}
\end{mybox}

\begin{mybox}{\texttt{r2c4}}
\footnotesize{\textbf{Z3 Proof} for \texttt{r2c4} shaded:}
    \begin{lstlisting}[basicstyle=\scriptsize\ttfamily]
(let ((a!1 (unit-resolution (asserted (or (not c02_04) (not c04_04)))
                            (asserted c04_04)
                            (not c02_04))))
  (unit-resolution a!1 (asserted c02_04) false))
    \end{lstlisting}

\footnotesize{\textbf{LLM Explanation:}  
Cell (2,4): \emph{Shaded}  
Reason:  
This cell contains the same symbol (4) as \texttt{(4,4)}, and both are in the same column.  
Since \texttt{(4,4)} has already been determined to be unshaded, \texttt{(2,4)} must be shaded to satisfy the Uniqueness Rule — no duplicate numbers unshaded in a column.}
\end{mybox}
To maintain neighbor connectivity, \texttt{r3c5} must be unshaded.

\begin{mybox}{\texttt{r3c5}}
\footnotesize{\textbf{Z3 Proof} for \texttt{r3c5} unshaded:}
    \begin{lstlisting}[basicstyle=\scriptsize\ttfamily]
(unit-resolution (unit-resolution (asserted (or c01_05 c02_04 c03_05))
                                  (asserted (not c01_05))
                                  (asserted (not c02_04))
                                  c03_05)
                 (asserted (not c03_05)) false)
    \end{lstlisting}

\footnotesize{\textbf{LLM Explanation:}  
Cell (3,5): \emph{Unshaded}  
Reason:  
To satisfy the Unshaded Neighbor Lemma, cell \texttt{(2,5)} must have at least one unshaded neighbor. But:  
\texttt{(1,5)} is shaded,  
\texttt{(2,4)} is shaded (as just established),  
So the only candidate left is \texttt{(3,5)}.  
To preserve the rule that every cell has at least one unshaded neighbor, \texttt{(3,5)} must be unshaded.}
\end{mybox}

Finally, we address the remaining dependencies around column 1. The following inferences determine the status of \texttt{r4c2} and eventually resolve \texttt{r5c1}.

\begin{figure}[t]
  \centering
  \begin{minipage}[b]{0.32\textwidth}
    \centering
    \includegraphics[width=\textwidth]{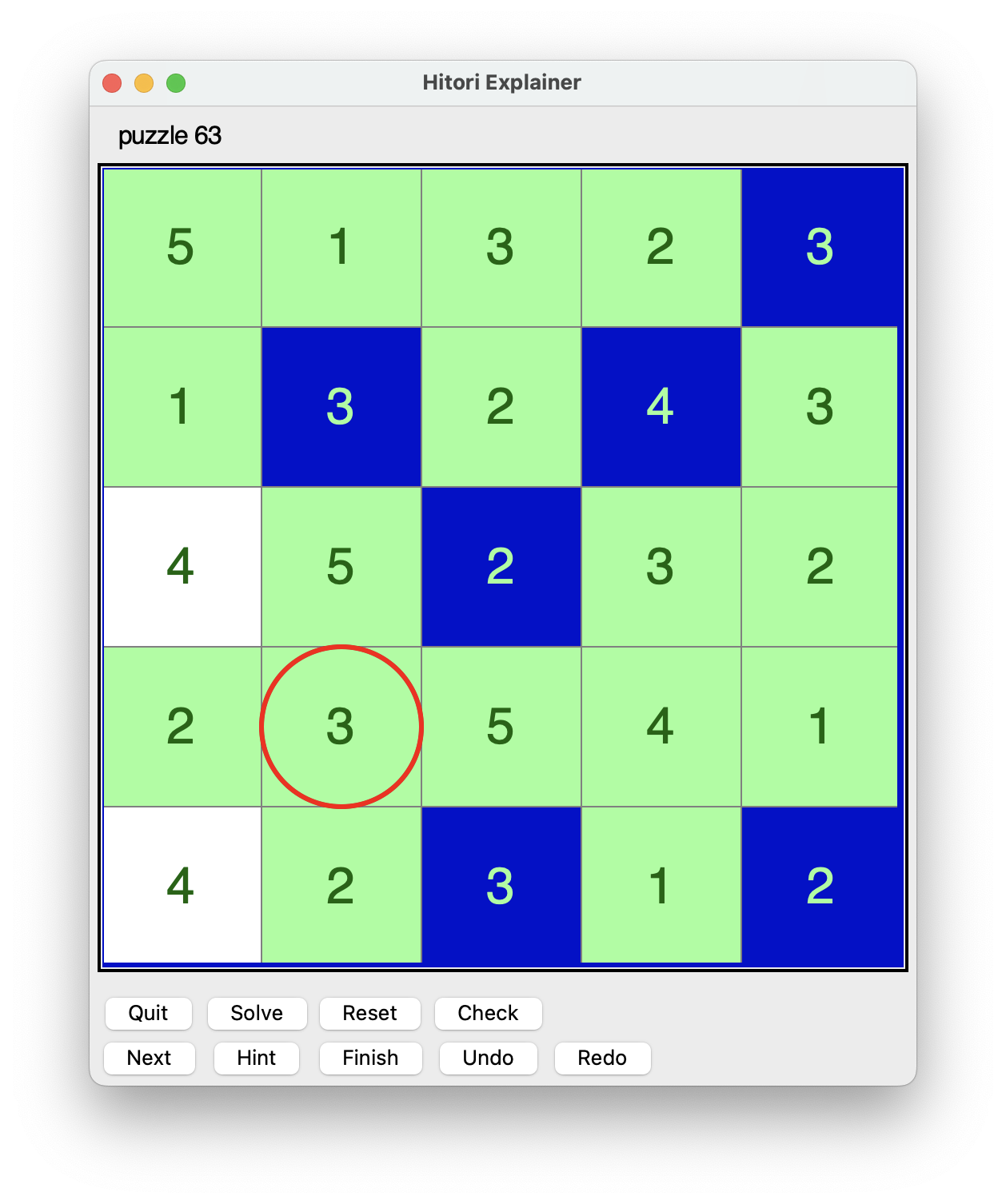}
    \caption*{(a) Deciding \texttt{r4c2}.}
  \end{minipage}
  \hfill
  \begin{minipage}[b]{0.32\textwidth}
    \centering
    \includegraphics[width=\textwidth]{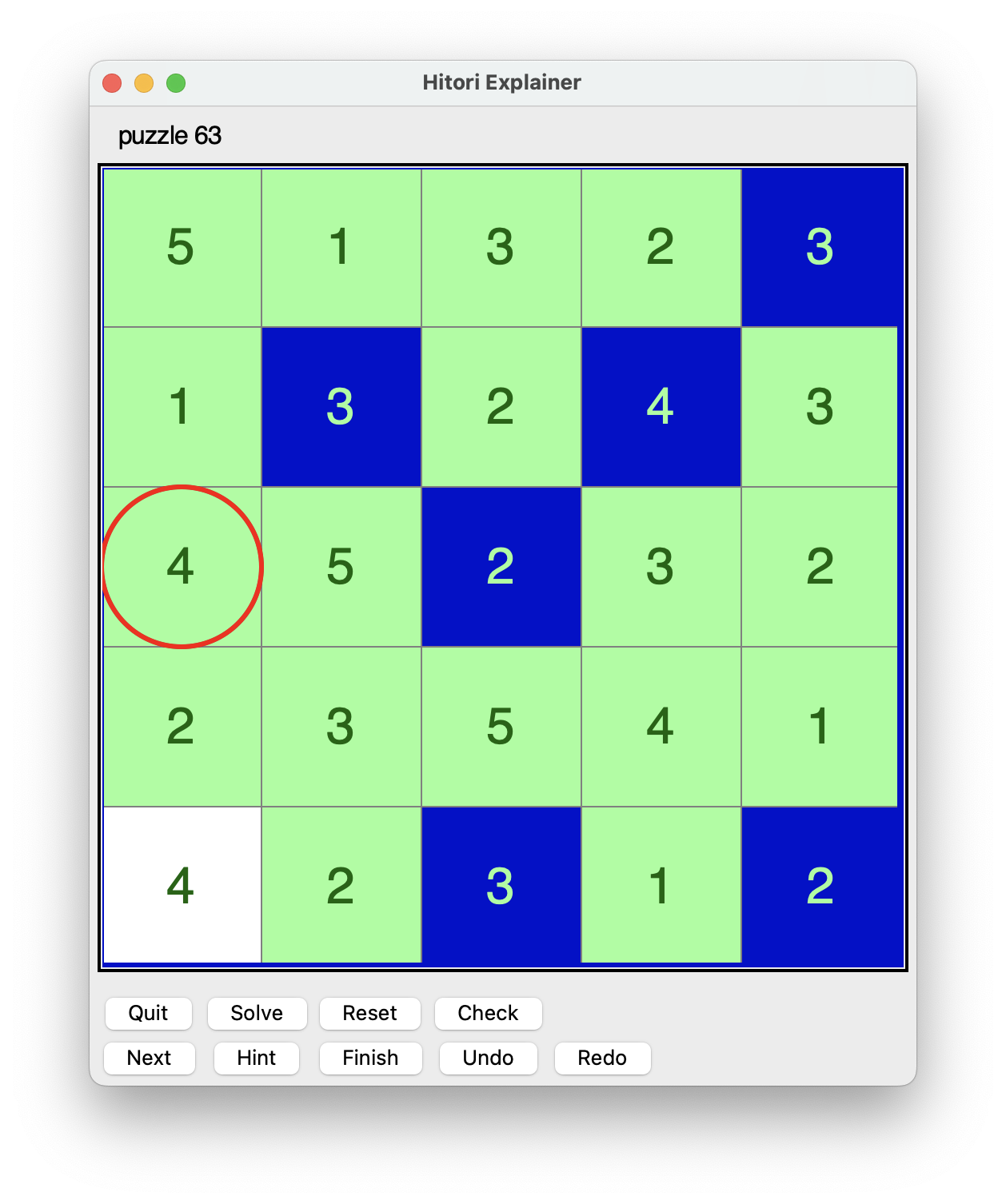}
    \caption*{(b) Deciding Cell \texttt{r3c1}}
  \end{minipage}
  \hfill
  \begin{minipage}[b]{0.32\textwidth}
    \centering
    \includegraphics[width=\textwidth]{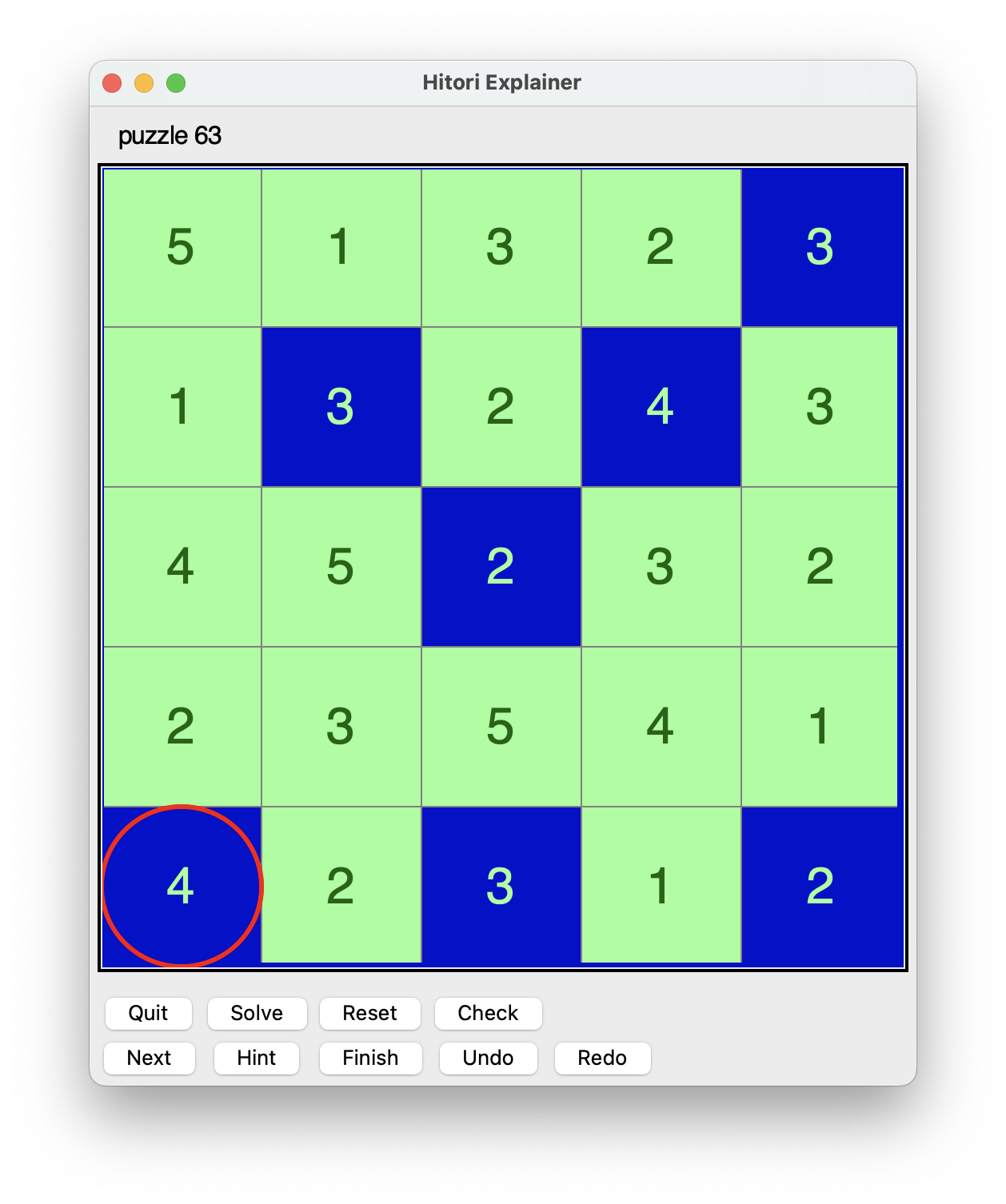}
    \caption*{(c) Deciding Cell \texttt{r5c1}}
  \end{minipage}
  \caption{Stepwise justification of key shading and unshading decisions in the Hitori puzzle (4).}
  \label{fig:three-images-four}
\end{figure}

\begin{mybox}{\texttt{r4c2}}
\footnotesize{\textbf{Z3 Proof} for \texttt{r4c2} unshaded:}
    \begin{lstlisting}[basicstyle=\scriptsize\ttfamily]
(let ((a!1 (unit-resolution (unit-resolution (asserted (or c04_02 c05_01 c05_03))
                                             (asserted (not c05_03))
                                             (or c04_02 c05_01))
                            (asserted (not c04_02))
                            c05_01))
      (a!2 (unit-resolution (unit-resolution (asserted (or c02_02
                                                           c03_01
                                                           c03_03
                                                           c04_02))
                                             (asserted (not c02_02))
                                             (asserted (not c03_03))
                                             (or c03_01 c04_02))
                            (asserted (not c04_02))
                            c03_01)))
  (unit-resolution (asserted (or (not c03_01) (not c05_01))) a!1 a!2 false))
    \end{lstlisting}

\footnotesize{\textbf{LLM Explanation:}  
Cell (4,2): \emph{Unshaded}  
Reason:  
Let’s unpack the chain of reasoning that forces \texttt{(4,2)} to be unshaded:

\textbf{Context:}  
The goal is to avoid violating the Unshaded Neighbor Lemma and the Uniqueness Rule.  
We’re assuming \texttt{(4,2)} is shaded and showing this leads to a contradiction.

\begin{itemize}
    \item \textbf{Step 1: Impact on (5,2)’s neighborhood}  
    If \texttt{(4,2)} is shaded, then in order to satisfy the Unshaded Neighbor Lemma at \texttt{(5,2)},  
    either \texttt{(5,1)} or \texttt{(5,3)} must be unshaded.  
    But \texttt{(5,3)} is already known to be shaded, so \texttt{(5,1)} must be unshaded.

    \item \textbf{Step 2: Impact on (3,2)’s neighborhood}  
    Still assuming \texttt{(4,2)} is shaded, to ensure \texttt{(3,2)} has an unshaded neighbor, one of the following must be unshaded:  
    \texttt{(2,2)} (already known to be shaded),  
    \texttt{(3,1)},  
    \texttt{(3,3)} (already shaded),  
    or \texttt{(4,2)} (assumed shaded).  
    This forces \texttt{(3,1)} to be unshaded.

    \item \textbf{Step 3: Conflict in column 1}  
    Now we’ve concluded:  
    \texttt{(5,1)} is unshaded (from Step 1),  
    \texttt{(3,1)} is unshaded (from Step 2),  
    But they both contain the same number (4) and lie in the same column.  
    This violates the Uniqueness Rule, which forbids duplicate unshaded symbols in a column.  
    Contradiction.
\end{itemize}

\textbf{Conclusion:}  
Assuming \texttt{(4,2)} is shaded leads to a contradiction involving neighbor requirements and column uniqueness.  
Therefore, \texttt{(4,2)} must be unshaded.}
\end{mybox}

A different explanation for \texttt{r4c2} would note that shading it disconnects the left and right sides of the grid.  Our program, however, currently prefers weak proofs like the one above, whenever they are available.

\begin{mybox}{\texttt{r5c1}}
\footnotesize{\textbf{Z3 Proof} for \texttt{r5c1} shaded:}
    \begin{lstlisting}[basicstyle=\scriptsize\ttfamily]
(let ((a!1 (unit-resolution (asserted (or (not c03_01) (not c05_01)))
                            (asserted c03_01)
                            (not c05_01))))
  (unit-resolution a!1 (asserted c05_01) false))
    \end{lstlisting}

\footnotesize{\textbf{LLM Explanation:}  
Cell (5,1): \emph{Shaded}  
Reason:  
Cells \texttt{(3,1)} and \texttt{(5,1)} are in the same column and both contain the same number (4).  
Since \texttt{(3,1)} is already confirmed as unshaded, \texttt{(5,1)} must be shaded to satisfy the Uniqueness Rule.}
\end{mybox}

\section{Futoshiki}

We give here an example of the promising preliminary results for puzzle genres other than Hitori that were mentioned in Section~\ref{sec:limitations}.  Futoshiki is a number-placement puzzle played on an $n \times n$ grid that initially contains some (given) numbers and some inequality signs.  The objective is to complete a Latin square with the numbers $1$ to $n$ so that the inequalities are satisfied.  As usual with pencil puzzles, a well-formed Futoshiki puzzle has a unique solution.  Deciding whether a Futoshiki puzzle has a solution is NP-complete (by reduction from completion of partial Latin squares).  Below is a small ($4 \times 4$) puzzle alongside its solution.
\begin{center}
    \begin{tikzpicture}[scale=0.7,transform shape]
    \sffamily\sansmath\LARGE
    \begin{scope}
      \fill[yellow!5!white] (0,0) rectangle (4,4);
      \draw[step=1cm,ultra thin] (0,0) grid (4,4);
      \foreach \i in {1,...,4} {
        \node[gray] at (-0.5,4.5-\i) {\Large\i};
      }
      \foreach \j in {1,...,4} {
        \node[gray] at (\j-0.5,4.5) {\Large\j};
      }
      \node[purple, rotate=180, anchor=center] at (1.0, 3.5) {$<$};
      \node[purple, rotate=-90, anchor=center] at (0.5, 2.0) {$<$};
      \node[purple, rotate=-90, anchor=center] at (1.5, 2.0) {$<$};
      \node[purple, rotate=90, anchor=center] at (3.5, 2.0) {$<$};
      \node[purple, rotate=-90, anchor=center] at (3.5, 1.0) {$<$};
      \node[purple, rotate=90, anchor=center] at (2.5, 1.0) {$<$};
      \node[purple] at (2.5,3.5) {1};
      \node[purple] at (3.5,3.5) {3};
      \draw[very thick] (0,0) rectangle (4,4);
    \end{scope}
    \begin{scope}[xshift=6cm]
      \fill[yellow!5!white] (0,0) rectangle (4,4);
      \draw[step=1cm,ultra thin] (0,0) grid (4,4);
      \foreach \i in {1,...,4} {
        \node[gray] at (-0.5,4.5-\i) {\Large\i};
      }
      \foreach \j in {1,...,4} {
        \node[gray] at (\j-0.5,4.5) {\Large\j};
      }
      \node[purple, rotate=180, anchor=center] at (1.0, 3.5) {$<$};
      \node[purple, rotate=-90, anchor=center] at (0.5, 2.0) {$<$};
      \node[purple, rotate=-90, anchor=center] at (1.5, 2.0) {$<$};
      \node[purple, rotate=90, anchor=center] at (3.5, 2.0) {$<$};
      \node[purple, rotate=-90, anchor=center] at (3.5, 1.0) {$<$};
      \node[purple, rotate=90, anchor=center] at (2.5, 1.0) {$<$};
      \node[black] at (0.5,3.5) {4};
      \node[black] at (1.5,3.5) {2};
      \node[purple] at (2.5,3.5) {1};
      \node[purple] at (3.5,3.5) {3};
      \node[black] at (0.5,2.5) {1};
      \node[black] at (1.5,2.5) {3};
      \node[black] at (2.5,2.5) {4};
      \node[black] at (3.5,2.5) {2};
      \node[black] at (0.5,1.5) {2};
      \node[black] at (1.5,1.5) {4};
      \node[black] at (2.5,1.5) {3};
      \node[black] at (3.5,1.5) {1};
      \node[black] at (0.5,0.5) {3};
      \node[black] at (1.5,0.5) {1};
      \node[black] at (2.5,0.5) {2};
      \node[black] at (3.5,0.5) {4};
      \draw[very thick] (0,0) rectangle (4,4);
    \end{scope}
  \end{tikzpicture}
\end{center}

The easier starting point for the solution of this puzzle is Row~$1$.  Only $2$ and $4$ remain to be placed, and the inequality determines their positions.  It is also easy to see that, in Column~$4$, the $1$ must go in Row~$3$.  The next step is the key one.  One observes that anything but a 2 in \texttt{r3c1} would force three numbers that are at least $3$ in Row~$3$.  The resolution proof produced by Z3 for this step and its interpretation by ChatGPT o4 are shown below.

\begin{mybox}{\texttt{r3c1 = 2}}
\footnotesize{\textbf{Z3 Proof} for \texttt{r3c1 = 2}:}
\begin{lstlisting}[basicstyle=\scriptsize\ttfamily]
(let ((a!1 (unit-resolution (asserted (or (not x_1_1_4) (not x_3_1_4)))
             (asserted x_1_1_4)
             (not x_3_1_4)))
      (a!3 (unit-resolution (asserted (or x_3_1_3 (not x_3_1_2)))
             (asserted x_3_1_2)
             x_3_1_3))
      (a!4 (unit-resolution (asserted (or x_3_3_3 (not x_4_3_2)))
             (unit-resolution (asserted (or x_1_3_2 x_4_3_2))
                              (asserted (not x_1_3_2))
                              x_4_3_2)
             x_3_3_3))
      (a!5 (unit-resolution (asserted (or x_1_2_3 x_3_2_3 (not x_1_2_2) (not x_3_2_2)))
             (asserted x_3_2_2)
             (asserted x_1_2_2)
             (asserted (not x_1_2_3))
             x_3_2_3)))
  (let ((a!2 (unit-resolution (asserted (or x_3_1_4 x_3_3_4 (not x_3_1_3) (not x_3_3_3)))
               a!1
               (or x_3_3_4 (not x_3_1_3) (not x_3_3_3))))
        (a!6 (unit-resolution (asserted (or x_3_1_4 x_3_2_4 (not x_3_1_3) (not x_3_2_3)))
               a!1
               a!5
               (or x_3_2_4 (not x_3_1_3)))))
    (unit-resolution (asserted (or (not x_3_2_4) (not x_3_3_4)))
                     (unit-resolution a!2 a!3 a!4 x_3_3_4)
                     (unit-resolution a!6 a!3 x_3_2_4)
                     false)))
\end{lstlisting}

\footnotesize{\textbf{LLM Explanation:}  
This proof shows that \texttt{r3c1} must be exactly 2. Here's why:
\begin{itemize}
    \item \texttt{r1c1} is already 4 (given), so \texttt{r3c1} cannot also be 4 due to column uniqueness.
    \item Suppose \texttt{r3c1} is 3. Then \texttt{r2c1} must be less than 3 (to satisfy the inequality \texttt{r2c1 < r3c1}), so say \texttt{r2c1} = 1 or 2.
    \item But now, due to the inequality \texttt{r3c2 > r2c2}, \texttt{r3c2} must be at least 3. Meanwhile, \texttt{r1c2} is 2 (given), so \texttt{r3c2} must be greater than 2 to avoid duplication in the column. So \texttt{r3c2} must be at least 3.
    \item These constraints together force both \texttt{r3c2} and \texttt{r3c3} to be at least 4.
    \item But that would place the number 4 in both \texttt{r3c2} and \texttt{r3c3}, violating the Futoshiki rule that each number must appear exactly once in a row.
\end{itemize}}

Thus, assuming \texttt{r3c1 = 3} leads to a contradiction. Since values 1 and 4 are already excluded for \texttt{r3c1}, it must be 2.

\end{mybox}
The prompt used to obtain this explanation included, beside the resolution proof, an explanation of the encoding, stating that \texttt{x\_i\_j\_k} is true if, and only if, \texttt{ricj} is at least $k$, and a few clauses annotated with their role in the encoding.  For example, the prompt explains that \texttt{(or x\_3\_1\_4 x\_3\_3\_4 (not x\_3\_1\_3) (not x\_3\_3\_3))} encodes the constraint that \texttt{r3c1} and \texttt{r3c3} cannot both be $3$.  This puzzle is simple enough that the LLM could solve it without recourse to the Z3 proofs.  However, it could not satisfactorily explain the solution by itself.

\end{document}